\pdfoutput=1

\documentclass[11pt]{article}

\usepackage{emnlp2022}

\usepackage{times}
\usepackage{latexsym}

\usepackage[T1]{fontenc}

\usepackage[utf8]{inputenc}

\usepackage{microtype}

\usepackage{csquotes}
\usepackage{xspace}
\usepackage{graphicx}
\usepackage{booktabs}
\usepackage{multirow}
\usepackage{paralist}
\usepackage{enumitem}

\newcommand{\draftonly}[1]{#1}

\newcommand{\draftcomment}[1]{\draftonly{#1}}
\newcommand{\swabha}[1]{\draftcomment{\textcolor{orange}{\small [#1]$_{{SS}}$}}}
\newcommand{\maxma}[1]{\draftcomment{\textcolor{red}{\small [#1]$_{\textrm{max}}$}}}
\newcommand{\jiao}[1]{\draftcomment{\textcolor{magenta}{\small [#1]$_{\textrm{jiao}}$}}}
\usepackage{arydshln}

\makeatletter
\def\adl@drawiv#1#2#3{%
        \hskip.5\tabcolsep
        \xleaders#3{#2.5\@tempdimb #1{1}#2.5\@tempdimb}%
                #2\z@ plus1fil minus1fil\relax
        \hskip.5\tabcolsep}
\newcommand{\cdashlinelr}[1]{%
  \noalign{\vskip\aboverulesep
           \global\let\@dashdrawstore\adl@draw
           \global\let\adl@draw\adl@drawiv}
  \cdashline{#1}
  \noalign{\global\let\adl@draw\@dashdrawstore
           \vskip\belowrulesep}}
\makeatother

\newcommand{\RNum}[1]{\uppercase\expandafter{\romannumeral #1\relax}}

\newcommand{\mkclean}{
	\renewcommand{\swabha}[1]{}
	\renewcommand{\maxma}[1]{}
	\renewcommand{\jiao}[1]{}
}

\mkclean

\newcommand{\cose}{{\fontfamily{cmss}\selectfont{CoS-E}}\xspace}
\newcommand{\ecqa}{{\fontfamily{cmss}\selectfont{ECQA}}\xspace}
\newcommand{\quartz}{{\fontfamily{cmss}\selectfont{QuaRTz}}\xspace}

\title{Investigating the Benefits of Free-Form Rationales}

\author{Jiao Sun$^1$ \, Swabha Swayamdipta$^1$ \, Jonathan May$^{1,2}$\thanks{\xspace\xspace Work done prior to JM joining Amazon.} \, Xuezhe Ma$^{1,2}$\\
$^1$University of Southern California \\
$^2$Information Sciences Institute \\
\texttt{\{jiaosun,swabhas\}@usc.edu \, \{jonmay,xuezhema\}@isi.edu}}

\begin{document}
\maketitle

\begin{abstract}
Free-form rationales aim to aid model interpretability by supplying background knowledge that can help understand model decisions.
Popular commonsense QA datasets such as \cose and \ecqa provide crowdsourced free-form rationales for instances, but their utility remains under-investigated. 
We present studies which show that 88\% of \ecqa rationales indeed provide humans additional background information to understand a decision, while 93\% of \cose rationales do not. 
Inspired by this finding, we ask: can the additional context provided by free-form rationales benefit models, similar to their effect on human users? 
We investigate the usefulness of rationales as an additional training signal, by varying the quantity and quality of rationales during training. 
After controlling for instances where rationales leak the correct answer while not providing additional background knowledge, we find that incorporating only 5\% of rationales during training can boost model performance by 47.22\% for \cose and 57.14\% for \ecqa during inference.
Moreover, we also show that rationale quality matters: compared to crowdsourced rationales, T5-generated rationales provide not only a weaker training signal, but are also not helpful for humans in aiding model interpretability.
\end{abstract}

\section{Introduction}
\label{sec:intro}

Free-form rationales designed to explain decisions by providing additional world knowledge or commonsense reasoning, are key for interpretability \cite{kim2015interactive,lipton2018mythos,AlvarezMelis2018TowardsRI} in natural language processing tasks.\footnote{
We use the terms ``rationale'' and ``explanation'' interchangeably. 
Please see \citet{Wiegreffe2021TeachMT} and \citet{jacovi-goldberg-2021-aligning} for more details on terminology.}
Free-form rationales come with the promise of being easily interpretable by humans, in contrast to other kinds of explanations, such as extractive rationales in the form of textual highlights \cite{Camburu2018eSNLINL,lei-etal-2016-rationalizing}, or low-level neuron activations in neural architectures~\cite{Hohman2020SummitSD}. 
Indeed, there have been increasing efforts to collect corpora containing free-form rationales for task instances, which provide a supervised setting for teaching models to produce rationales for test-time decisions. 
Such corpora include \cose~\cite{rajani-etal-2019-explain} and \ecqa~\cite{aggarwal-etal-2021-explanations} for commonsense question-answering, e-SNLI \cite{Camburu2018eSNLINL} for natural language inference, SBIC \cite{sap-etal-2020-social} for social bias inference, among others. 

\begin{figure}[t]
    \centering
    \includegraphics[width=\linewidth]{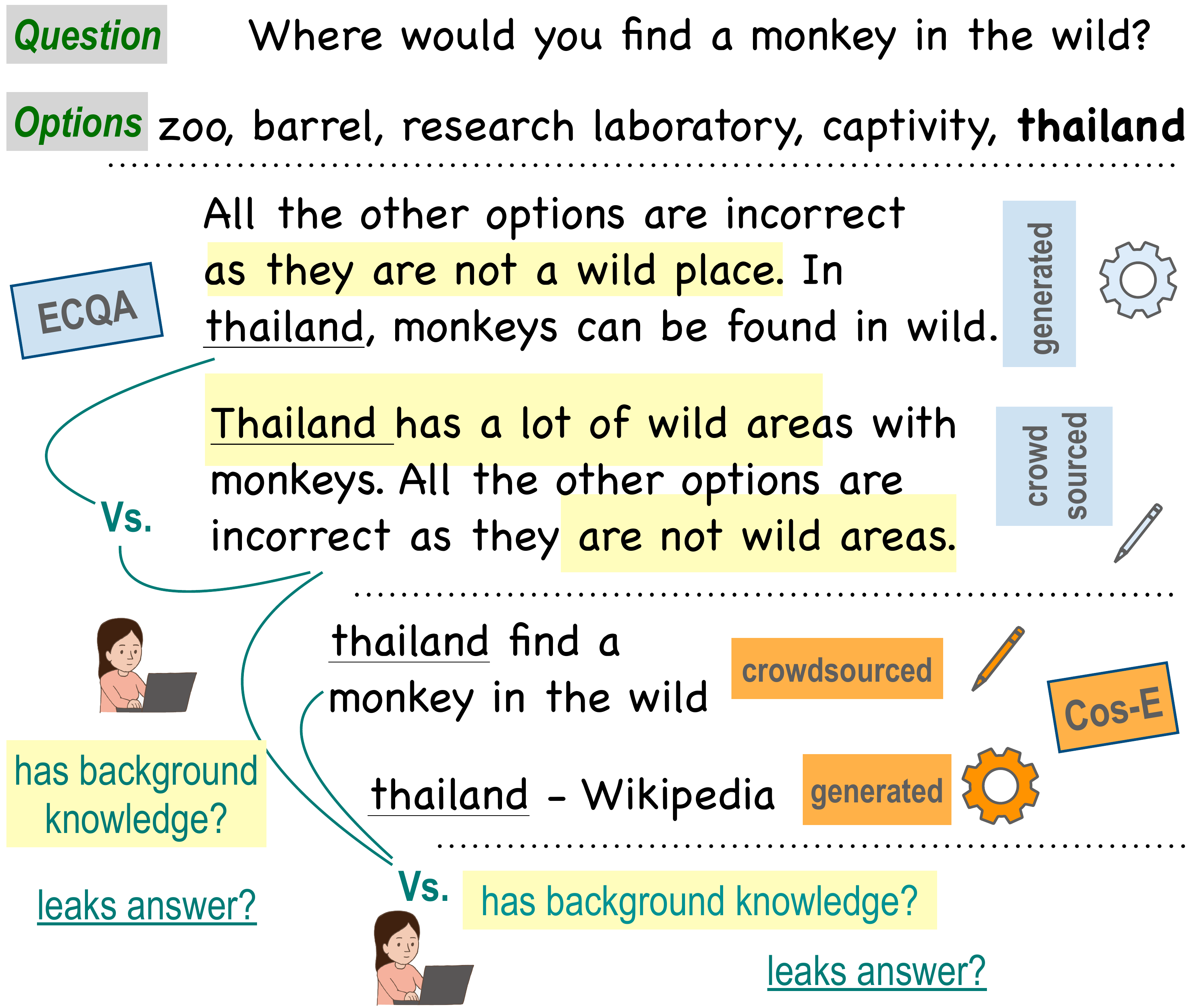}
    \caption{
    Illustration of our investigation into free-form rationales for commonsense QA from \cose \cite{trajanovski-etal-2021-text} and \ecqa \cite{aggarwal-etal-2021-explanations}.
    We conduct \emph{human} studies to understand perceived usefulness of rationales, by asking if they contain background knowledge necessary to answer a question (yellow highlights).
    We also investigate if rationales leak the answer to \emph{models} that use them as additional training signals. 
    Our work compare rationales from different sources, and finds that \ecqa rationales are preferable to \cose rationales on various axes.
    Finally, we find that crowdsourced rationales also offer greater benefits to both humans and models than generated rationales.
    }
    \label{fig:teaser}
\end{figure}


However, the benefits of rationales remain unclear. 
Do crowdsourced rationales really help human users interpret decisions better, or do they simply provide the right answer without the necessary background knowledge or reasoning? 
Our work explores this question through two carefully designed human studies.
We find that rationales from different corpora have different capabilities: humans find 93\% of \ecqa rationales provide additional information that can help answer questions, while only 12\% of \cose rationales do. 

Inspired by this finding, we further ask: analogous to the benefit to human users, can crowdsourced rationales \textit{also benefit models} by providing an additional training signal to boost performance?
In contrast to prior work that uses rationales as supervision to generate model rationales, we focus on using crowdsourced rationales to simply aid a task model's classification capabilities.
Our results indicate that while crowdsourced rationales do indeed boost model performance, they might be doing so trivially, i.e. by simply leaking the correct answer to the model.
In response, we experiment with different strategies for altering \ecqa and \cose rationales to prevent such leakage, and set up a fair test benchmark.
We find that, even without leakage, rationales with background knowledge are helpful: including only 5\% of high-quality rationales during training can improve model performance by 47.22\% at inference time. 
This finding generalizes to \quartz~\cite{tafjord-etal-2019-quartz}, a dataset for textual relationship inference, in which rationales are designed explicitly to not leak the ground truth.

Finally, we investigate if \textit{automatically generated} rationales provide similar benefits as crowdsourced rationales.
Our human studies indicate that the perceived usefulness of generated rationales from T5 \cite{t5} is much lower than that of human-written ones.
Moreover, we find that these generated rationales are not particularly valuable as training signals. 
Overall, our results indicate that the quality of free-form rationales in existing popular datasets is paramount for both human interpretability as well as model supervision.\footnote{\url{https://github.com/sunjiao123sun/rationale-utility}.}

\section{Preliminaries}
\label{sec:tasks}

\paragraph{Tasks and Datasets.} 
We explore three large datasets containing crowdsourced free-form natural language rationales.
The first two, \cose~\cite{rajani-etal-2019-explain} and \ecqa \cite{aggarwal-etal-2021-explanations}, address commonsense-based question answering (ComQA). 
The ComQA task is based on answering questions about common situations, from a choice of 3 (\cose v1.0) or 5 (\cose v1.11) answers, along with providing a free-text explanation for the correct answer.\footnote{
\cose does not provide explanations for instances in the test set; we report our results on its validation set. 
}
\ecqa builds upon and improves the quality of \cose v1.11 explanations, in terms of comprehensiveness, refutation completeness and non-redundancy~\cite{aggarwal-etal-2021-explanations}. 
In addition, \ecqa explanations are contrastive, i.e. they include rationales for choosing the correct option and rejecting other options (see Tables~\ref{tab:cose-ecqa-shuffle}, \ref{tab:human-indirect-example}, and Table~\ref{tab:qualitative} in Appendix~\ref{app:generation-example} for examples). 
\begin{table}[t]
\small
\centering
\begin{tabular}{p{0.12\linewidth}|p{0.75\linewidth}}
\toprule
\textbf{Source}       & \textbf{Rationale}               \\ \midrule
\cose v1.11  & People waiting alongside with when you're in a reception area                                                                                                                                                      \\ \midrule
\ecqa       & People waits in a reception area. You cant wait along with a motel, hotel, chair or a hospital. These are the people where the reception area is found but people waits together at reception area of such places. \\ \midrule
\ecqa-shuffle & You cant wait along with a motel, hotel, chair or a hospital. These are the people where the reception area is found but people waits together at reception area of such places. People waits in a reception area. \\ \bottomrule
\end{tabular}
\caption{Example annotations from \cose v1.11 and \ecqa for the question \emph{``What are you waiting alongside with when you're in a reception area?''} with options \emph{1: motel 2: chair 3: hospital 4: people 5: hotels} and the correct option \emph{people}. \cose annotation directly combines the question and the correct answer, while \ecqa annotation provides additional background knowledge.}
\label{tab:cose-ecqa-shuffle}
\end{table}

We additionally consider an open-domain reasoning task about textual qualitative relationships, via the \quartz~\cite{tafjord-etal-2019-quartz} dataset, for a subset of our experiments.
In this task, each instance contains a triplet: a situated qualitative question, two answer options and a knowledge statement that can help  answer the question. 
For example, for \emph{``Compared to a box of bricks a box of feathers would be (A) lighter (B) heavier''}, the annotated knowledge in \quartz is \emph{A given volume of a denser substance is heavier than the same volume of a less dense substance}. 
In contrast to \cose and \ecqa, the two options for a question in \quartz are orthogonal, which means the knowledge provided to support one option will automatically reject the other option. 
Furthermore, this general qualitative knowledge statement in \quartz is guaranteed to not leak the correct answer. 
While not explicitly designed for interpretability, we treat the annotated knowledge in \quartz as a rationale that can help understand or derive the correct answer. 
See dataset stats in Table~\ref{tab:statistics} in App~\ref{app:t5}.

\section{Do crowdsourced rationales aid human interpretability?}
\label{sec:human}

Free-text rationales purportedly improve human user interpretability by explaining a model's decisions in natural language. 
We seek to discover which characteristics of the rationales aid users:
\begin{itemize}
    \item [Q1] Do rationales provide \textbf{additional background knowledge} for understanding decisions? E.g., the rationale: `\emph{Air cannot stay in any object that has a hole in it}' provides additional knowledge for understanding why the answer to `\emph{What would not be true about a basketball if it had a hole in it but it did not lose its general shape?}'  should be `\emph{full of air}'.
    \item [Q2] Do rationales provide explicit clues to \textbf{leak} the correct answer? For ComQA, this might initially seem like a helpful rationale, without really being so.\footnote{
    While leakage does not reduce the utility of a rationale for human interpretability, it does have implications for utility as model supervision, as we will see in subsequent sections, \S\ref{sec:experiments} and \S\ref{sec:generated}.
    }
    E.g., given a rationale: `\emph{Mexico is one of the largest coffee production country.}', one can guess the correct answer should be `\textit{mexico}', when given the options `\emph{mildred’s coffee shop}',  `\emph{mexico}', `\emph{diner}', \emph{kitchen}' or `\emph{canteen}', \underline{without} looking at the question `\emph{In what Spanish speaking North American country can you get a great cup of coffee?}'. 
\end{itemize}

\subsection{Preliminary Studies}
\label{sec:studies}

We investigate Q1 and Q2 via a direct assessment (\S\ref{sec:human-direct}) by human raters, as well as via proxy questions offering an indirect assessment (\S\ref{sec:human-proxy}) by the raters.

\subsubsection{Direct Assessment}
\label{sec:human-direct}

We conduct a {pilot study} where given the question, options, correct answer and rationales from \cose and \ecqa for a ComQA instance, annotators are tasked to \textit{directly} answer which rationale provides \textit{additional background knowledge} that can help them answer the question. 
Four options are possible: \cose, \ecqa, \emph{neither}, or \emph{both}.\footnote{While \citet{aggarwal-etal-2021-explanations} provide similar human studies comparing \ecqa and \cose rationales, they do not specifically ask for additional background knowledge.} Simultaneously, we ask annotators if any of the two rationales leaks the correct answer. 
Concretely, the annotators are required to provide three annotations for each instance:
\begin{itemize}
    \item  \emph{choose one option for the additional background information} (T1);
    \item \emph{judge if} \ecqa \emph{rationale leaks the correct answer} (T2);
    \item \emph{judge if} \cose \emph{leaks the correct answer} (T3).
\end{itemize}


We conduct our study on the first 120 rationales in \ecqa and \cose v1.11 test set via the Amazon Mechanical Turk platform. 
For each instance, we collect annotations from three independent annotators.
Using Fleiss's Kappa \cite{Fleiss1973TheEO}, the inter annotator agreement (IAA) for T1, T2 and T3 are 0.43, 0.26, and 0.30, respectively, indicating moderate agreement. 
We take the majority vote as the final label.\footnote{Further details on this study are in the Appendix~\ref{app:anno-human-study}, including Fig.~\ref{fig:human-direct} showing our annotation interface.}

Table~\ref{tab:human-direct} shows the results of our human evaluation. 
We see 85.8\% of \ecqa rationales provide additional background knowledge to help answer the question, while only 30.0\% of \cose rationales do the same, indicating greater usefulness of \ecqa rationales for human interpretability.
Both \ecqa and \cose rationales leak the correct answers. 
Indeed, most \ecqa rationales provide some background knowledge necessary for humans to understand the decision, while also revealing the correct answer; the same does not hold for \cose.
\begin{table}[t!]
\centering
\small
\resizebox{0.5\textwidth}{!}{
\begin{tabular}{@{}lrrrr@{}}
\toprule
\textbf{} & \textbf{$\mathbf{R}_\textrm{\ecqa}$} & \textbf{$\mathbf{R}_\textrm{\cose}$} & \textbf{both} & \textbf{neither}\\ \midrule
Q1: has bg. knowl.? & 65.0\% & 9.2\% & 20.8\% & 5.0\%\\ 
Q2: leaks answer? & 83.3\% & 43.3\% & n/a & n/a\\ \bottomrule
\end{tabular}
}
\caption{
Human study directly comparing \ecqa and \cose rationales on 120 ComQA instances, for the presence of background knowledge, and answer leakage.
}
\label{tab:human-direct}
\end{table}

\subsubsection{Indirect Assessment}
\label{sec:human-proxy}

While the previous study asked participants to directly assess the background knowledge of individual rationales, we design two other studies below that use a proxy to extract a human assessment of rationale utility \cite{tan2021diversity}, for Q2 and Q1, respectively.
Here, we randomly sample 100 ComQA instances from the test set.

\begin{table}[t]
\centering
\resizebox{0.5\textwidth}{!}{
\begin{tabular}{@{}lllll@{}}
\toprule
     & \textbf{question} & \textbf{options}& \textbf{$R^\textrm{crowd}$} & \textbf{$R^\textrm{constructed}$}  \\ 
     \midrule
\rotatebox{90}{\cose} &
  \begin{tabular}[c]{@{}l@{}}Where can\\a human\\find clothes\\that aren't\\ pants?\end{tabular} &
  \begin{tabular}[c]{@{}l@{}}pants shop,\\on planet earth,\\\textbf{dress shop},\\school,\\train wreck\end{tabular} &
  \begin{tabular}[c]{@{}l@{}}dress shop can a\\human find clothes\\that aren't pants. \end{tabular} &
  \begin{tabular}[c]{@{}l@{}}A human can find\\clothes at dress \\shop that aren't\\pants. \end{tabular}    \\
  \cdashlinelr{1-5}
\rotatebox{90}{\ecqa} &
 \begin{tabular}[c]{@{}l@{}}Where do\\adults use\\glue sticks?\end{tabular} &
  \begin{tabular}[c]{@{}l@{}}classroom,\\desk drawer,\\at school,\\\textbf{office},\\kitchen-\\ drawer\end{tabular} &
 \begin{tabular}[c]{@{}l@{}}Glue stick is a solid glue \\ used to stick thin paper \\ materials by adults in \\ offices. Adults don't go\\ to classroom and school,\\and other options don't \\ have adults.\end{tabular} &
  \begin{tabular}[c]{@{}l@{}}Adults use glue \\ sticks in their \\ offices. They do \\ not use them at \\ classroom, desk \\ drawer, at school \\ or kitchen drawer.\end{tabular}   \\
  \bottomrule
\end{tabular}
}
\caption{Examples of crowdsourced rationales for \cose and \ecqa, vs. our manually constructed rationales that declaratively combine the question and the answer without providing any background knowledge or commonsense reasoning. }
\label{tab:human-indirect-example}
\end{table}
\begin{table}[t]
\centering\small
\resizebox{0.5\textwidth}{!}{
\begin{tabular}{@{}lrrrr@{}}
\toprule
     & \textbf{$R^\textrm{crowd}$} & \textbf{$R^\textrm{constructed}$} & \textbf{neither}  & \textbf{either}\\
     \midrule
\cose & 3.0\%                & 5.0\%          & 92.0\%              & 0.0\%             \\
\ecqa & 73.0\%               & 9.0\%          & 14.0\%              & 4.0\%             \\ 
\bottomrule
\end{tabular}
}
\caption{
Results from our human study via indirect assessment to compare 100 pairs of crowdsourced and constructed rationales.
The IAA is 0.61.}
\label{tab:human-indirect-results}
\end{table}

For Q2, we ask annotators to guess the correct answer from all options, given only the crowdsourced rationales from \cose and \ecqa; annotators can also opt for ``cannot tell'' based on the evidence (see our interface in Appendix~\ref{app:anno-human-study}, Fig.~\ref{fig:human-indirect-leakage}).
We hypothesise that this study will indirectly answer whether the rationale leaks the correct option, if the worker is able to guess correctly.
Each instance is provided to three annotators, and we take a majority vote for their ratings.
We find that annotators are able to pick the correct answer, given only the rationales (and not questions) in 43.0\% of cases for \cose and 78.0\% of cases for \ecqa, with high agreement (IAA 0.73).
This confirms our findings from the direct assessment in Table~\ref{tab:human-direct}.

For Q1, 
we manually construct rationales to contrast with crowdsourced rationales. 
Our constructed rationales are designed to simply combine the question and the correct answer, but not provide any additional background knowledge.
If a human prefers the crowdsourced rationale, we can indirectly ascertain that it provides some background knowledge to help with human interpretability.
For \cose, we form a constructed rationale for a question by rephrasing the question as a statement and inserting the correct option in place of the question word. 
For \ecqa, in addition to the \cose-style constructed sentence, we add an additional sentence that rephrases the question as a negative statement, replaces some referents with pronoun anaphora, and inserts the incorrect options in place of the question word. 
We also try to ensure fluency and stylistic consistency with the crowdsourced explanations.

\begin{table*}[ht!]
\small
\resizebox{\textwidth}{!}{
\begin{tabular}{@{}lllr@{}}
\toprule
\textbf{Category} & \textbf{Description} & \textbf{Example}  & \textbf{Distribution} \\ 
\midrule
{\small\textbf{$R_{\textrm{no-leak-bg}}$}} 
& \begin{tabular}[c]{@{}l@{}}provides additional background \\knowledge without leaking \\ correct answers. \end{tabular} 
& \begin{tabular}[c]{@{}l@{}}\emph{Question:} What would not be true about a basketball if it had a hole \\ in it but it did not lose its general shape?  \\ \emph{Options:} 1: punctured 2: popular in america 3: \textbf{full of air} 4: gone 5: round \\ \emph{Rationale}: Air cannot stay in any object that has a hole in it.\end{tabular} 
& {\begin{tabular}[c]{@{}l@{}}4.83\%\\ (59/1221)\end{tabular}}\\ 
\cmidrule(lr){1-4}
{\small\textbf{$R_{\textrm{leak-bg}}$} }
& \begin{tabular}[c]{@{}l@{}}leaks the correct answer but contains  \\ additional background knowledge \\ that  can help answer questions.\end{tabular} 
& \begin{tabular}[c]{@{}l@{}}{\emph{Question:}} In what Spanish speaking North American country can you get a great cup of coffee?\\ \emph{Options:} 1: mildred's coffee shop 2: \textbf{mexico} 3: diner 4: kitchen 5: canteen \\ \emph{Rationale}: Mexico is one of the largest coffee production country.\end{tabular} 
& {\begin{tabular}[c]{@{}l@{}}6.72\%\\ (82/1221)\end{tabular}} \\ 
\cmidrule(lr){1-4}
{\small\textbf{$R_{\textrm{no-leak-no-bg}}$}}
& \begin{tabular}[c]{@{}l@{}}neither provides any additional \\ background information,  nor \\ leaks the correct answer.\end{tabular} 
& \begin{tabular}[c]{@{}l@{}}{\emph{Question:}} why would a person like to have a large house?\\ \emph{Options:} 1: have choice 2: mentally challenged 3: own house 4: obesity \textbf{5: lots of space} \\ \emph{Rationale}: This word is most relevant\end{tabular} 
& {\begin{tabular}[c]{@{}l@{}}43.65\%\\ (533/1221)\end{tabular}} \\ 
\cmidrule(lr){1-4}
{\small\textbf{$R_{\textrm{leak-no-bg}}$}}
& \begin{tabular}[c]{@{}l@{}}leaks the correct answer and does \\ not provide additional background \\knowledge.\end{tabular} 
& \begin{tabular}[c]{@{}l@{}}{\emph{Question:}} where will a cheap book be found?\\ \emph{Options:} 1: bookstore 2: classroom 3: \textbf{discount store} 4: school room 5: bedside table  \\ \emph{Rationale}:  discount shop retail shop\end{tabular} 
& {\begin{tabular}[c]{@{}l@{}}44.80\%\\ (547/1221)\end{tabular}} \\ 
\bottomrule
\end{tabular}
}
\caption{Our manual four-way categorization of \cose v1.11 (dev.) rationales, with examples.
Bolded options indicate ground truth. 
We find that 88.45\% of rationales do not provide additional background knowledge.
}
\label{tab:cose-categorization}
\end{table*}

We show two examples of our constructed rationales in Table~\ref{tab:human-indirect-example}. 
We provide human subjects with the question, the correct answer, the crowdsourced rationale (from \cose or \ecqa) and our constructed rationale. 
We instruct workers to choose the explanation that they would prefer if they need to explain the correct answer to someone who might not have the necessary background knowledge to understand given only the question and set of answer choices (see our interface in Appendix~\ref{app:anno-human-study}, Fig.~\ref{fig:human-indirect-background}).
Each instance is provided to three annotators, and we take a majority vote for their ratings.

Results in Table~\ref{tab:human-indirect-results} show that human raters overwhelmingly preferred neither our constructed rationales or \cose rationales, indicating that neither provides background knowledge necessary for answering the question.\footnote{Surprisingly, raters preferred the constructed rationales slightly over the crowdsourced rationales, which might be because some \cose rationales are off-topic; see Appendix~\ref{app:anno-human-study}.}
On the other hand, raters seem to prefer \ecqa rationales over our constructions, indicating that the former might contain background knowledge owing to their rigorous annotation procedure \cite{aggarwal-etal-2021-explanations}.
Yet, surprisingly, raters picked our constructed rationales 9\% of the time over \ecqa, while being ambivalent about either rationale for 4\% of the cases; moreover, they liked neither for 14\% of the cases!
This could indicate that some \ecqa instances might not provide adequate background knowledge, and / or raters might at times choose simpler (though vacuous) rationales; future work might pursue studying such cases.

\subsection{Categorizing Crowdsourced Rationales}
\label{sec:categorizing}

\paragraph{\textbf{\cose.}}
Although \citet{Narang2020WT5TT} criticize the quality of \cose rationales, 
\cose v1.11 is still widely used for commonsense reasoning \cite{paranjape-etal-2021-prompting}, analysis \cite{Majumder2021RationaleInspiredNL, wiegreffe-etal-2021-measuring}, and as an additional source of commonsense knowledge \cite{Ye2019AlignMA}.
In order for the community to understand the deficiencies of the crowdsourced \cose rationales, we provide a detailed study of the same, which was missing in \citet{Narang2020WT5TT}.

Building on Q1 and Q2, we aim to categorize \cose rationales into 4 categories, to determine if these provided background knowledge and/or leaked the answer. 
One of the authors manually categorized the rationales in the development set of \cose v1.11 into four categories.
To validate this categorization, three co-authors annotated a subset of 100 instances independently for the same categorization.
We obtained an IAA Fleiss Kappa of 0.65 for \emph{background knowledge} and 0.84 for \emph{leakage}, indicating moderate / high agreement. 
For these 100 instances, we use the majority vote among the three annotators as the final label. 
Appendix \ref{app:author} provides further details.

Table~\ref{tab:cose-categorization} describes and shows the distribution of the categories, with examples from each picked at random.
Rationales that do not provide additional background knowledge make up 88.45\% of the entire development set of \cose v1.11.
Using the development set as a lens, our annotation provides a qualitative and quantitative understanding of the crowdsourced \cose rationales. 
Future research should take into consideration these findings before using \cose rationales. 


\paragraph{\textbf{\ecqa.}}
\citet{aggarwal-etal-2021-explanations} build on \cose question-answer pairs and carefully collect detailed rationales. 
Table~\ref{tab:cose-ecqa-shuffle} compares \cose and \ecqa rationales, where the former directly combines the correct answer and the question, but the latter contains additional commonsense knowledge that can help answer the question, suggesting a higher quality. 
Moreover, \ecqa rationales are contrastive as they explain, for each option, why it is correct or incorrect.
Regardless, we find that all \ecqa rationales \textit{start} by explaining the correct option, followed by other options.
This ordering introduces a spurious correlation which likely provides a shortcut to a model for predicting the correct answer from the rationale, but for wrong reasons \cite{geirhos2020shortcut}.
A random shuffle of the sentences within each \ecqa rationale (last row; Table~\ref{tab:cose-ecqa-shuffle}) can address this issue.\footnote{We use the \href{https://spacy.io/}{Spacy} sentencizer to split the rationale, and randomly permute sentence ordering, with seed 0.}

\begin{table*}[ht!]
\resizebox{\textwidth}{!}{
\small
\begin{tabular}{@{}cccllllllll@{}}
\multicolumn{3}{r}{} & {$c1$} & {$c2$} & {$c3$} & {$c4$} & {$c5$} & {$c6$} & {$c7$} & {$c8$}\\ \toprule
 & \multirow{2}{*}{\rotatebox[origin=c]{90}{}} & \textbf{{Test}} $\rightarrow$ & {\textbf{I}} & {\textbf{I+$R_\textrm{\cose}$}}&
 \multicolumn{2}{c}{\textbf{I}+$R_\textrm{\ecqa}$} &
\multicolumn{4}{c}{\textbf{I+}$R_\textrm{\cose}$ (test subsets)} \\ 
\cmidrule(lr){4-4} \cmidrule(lr){5-5} \cmidrule(lr){6-7}  \cmidrule(lr){8-11}
 &  &{\textbf{\tiny{\%$R^\textrm{Tr}$}}} &  & & \multicolumn{1}{c}{\textbf{w/o shuffle}} & \multicolumn{1}{c}{\textbf{shuffled}} & \multicolumn{1}{c}{\textbf{$R_{\textrm{\small{no-leak-bg}}}$}} & \multicolumn{1}{c}{\textbf{$R_{\textrm{leak-bg}}$}} & \multicolumn{1}{c}{\textbf{$R_{\textrm{no-leak-no-bg}}$}} &
 \multicolumn{1}{c}{\textbf{$R_{\textrm{leak-no-bg}}$}}
 \\ \cmidrule(l){3-11}
{$r$1} & I$\rightarrow$O & \multicolumn{1}{r}{0\%} & 57.00 & 46.11 & 53.32 & 54.95 &  {40.68} &  {46.34} &  {45.97} &  {46.80}\\ \cmidrule(l){2-11} 
{$r$2} & \multirow{5}{*}{\rotatebox[origin=c]{90}{\tiny IR$^\textrm{Tr}_\textrm{\cose}\rightarrow$O}} & \multicolumn{1}{r}{5\%} & 53.78$_{1.10}$ & 72.53$_{2.19}$ & 76.50$_{2.30}$ & \multicolumn{1}{l|}{65.57$_{2.86}$} & 59.89$_{5.24}$ & 87.40$_{4.14}$ & 54.72$_{1.17}$ & 89.03$_{2.72}$ \\
{$r$3} & \multicolumn{1}{l}{} & \multicolumn{1}{r}{10\%} & 54.44$_{0.72}$ & 76.03$_{1.00}$ & 80.78$_{1.53}$ & \multicolumn{1}{l|}{63.74$_{0.78}$} & 70.06$_{2.88}$ &  89.02$_{3.45}$ & 56.85$_{1.48}$ & 93.42$_{0.65}$\\
{$r$4} & \multicolumn{1}{l}{} & \multicolumn{1}{r}{20\%} & 53.62$_{0.23}$ & 77.23$_{0.47}$ & 83.40$_{1.41}$ & \multicolumn{1}{l|}{62.71$_{1.80}$} & 68.93$_{5.59}$ & 95.53$_{1.15}$ & 56.97$_{1.08}$ & 95.12$_{0.09}$\\
{$r$5} & \multicolumn{1}{l}{} & \multicolumn{1}{r}{30\%} & 53.12$_{0.60}$ & 77.43$_{0.30}$ & 79.17$_{3.23}$ & \multicolumn{1}{l|}{63.56$_{1.28}$} & 73.55$_{5.46}$ & 94.71$_{0.58}$ & 56.72$_{1.11}$ & 96.49$_{0.67}$\\
{$r$6} & \multicolumn{1}{l}{} & \multicolumn{1}{r}{100\%} & 48.24 & 78.46 & 66.01 & \multicolumn{1}{l|}{64.46} & 71.19 & 97.56 & 57.97 & 96.34\\ \cmidrule(l){2-11} 
{$r$7} & \multirow{5}{*}{\rotatebox[origin=c]{90}{\tiny IR$^\textrm{Tr}_\textrm{\ecqa-shf.}\rightarrow$O}} & \multicolumn{1}{r}{5\%} & 54.05$_{0.95}$ & 59.27$_{0.91}$ & 86.65$_{1.10}$ & \multicolumn{1}{l|}{86.35$_{1.54}$} & 51.41$_{7.10}$ & 69.10$_{1.52}$ & 53.22$_{0.49}$ & 64.53$_{2.35}$\\
{$r$8} & \multicolumn{1}{l}{} & \multicolumn{1}{r}{10\%} & 54.05$_{1.08}$ & 61.72$_{2.11}$ & 92.55$_{0.52}$ & \multicolumn{1}{l|}{93.01$_{0.37}$} & 54.80$_{5.24}$ & 72.76$_{4.03}$ & 52.53$_{1.46}$ & 69.77$_{3.16}$\\
{$r$9} & \multicolumn{1}{l}{} & \multicolumn{1}{r}{20\%} & 53.29$_{0.32}$ & 66.50$_{0.66}$ & 95.41$_{0.48}$ & \multicolumn{1}{l|}{94.70$_{1.17}$} & 64.41$_{4.99}$ & 83.74$_{1.15}$ & 55.85$_{0.69}$ & 74.53$_{1.36}$\\
{$r$10} & \multicolumn{1}{l}{} & \multicolumn{1}{r}{30\%} & 52.85$_{0.67}$ & 65.05$_{0.78}$ & 95.85$_{0.34}$ & \multicolumn{1}{l|}{95.52$_{0.51}$} & 56.50$_{5.24}$ & 81.30$_{2.30}$ & 52.91$_{0.41}$ & 75.38$_{2.15}$\\
{$r$11} & \multicolumn{1}{l}{} & \multicolumn{1}{r}{100\%} & 38.08 & 67.32 & 97.3 & \multicolumn{1}{l|}{96.56} & 55.93 & 93.90 & 39.40 & 91.77 \\ \bottomrule
\end{tabular}
}
\caption{
ComQA accuracies under various train ({\textbf{\emph{r}}ow}) and test ({\textbf{\emph{c}}olumn}) settings. 
$r1$ is an I$\rightarrow$O T5 baseline without access to rationales during training; the following rows use different amounts ({\%R$^\textrm{Tr}$}) of \cose rationales ($r2-r6$) and shuffled \ecqa rationales ($r7-r11$) for training IR$\rightarrow$O T5 models.
At inference time, each model predicts the label given no rationale ($c1$), or given the crowdsourced rationales for the entire test set ($c2$-$c4$), or a subset of the \cose test set ($c5$-$c8$), selected based on the rationale categories in Table~\ref{tab:cose-categorization}.
$c4$ and $c3$ report \ecqa test set performance, when the test rationales are shuffled or not, respectively.
We report accuracies averaged across 3 random seeds (stdev as subscript) for \%R selection during training.
}
\label{tab:train-crowdR-comqa}
\end{table*}
\begin{table}[t]
\small
\centering
\resizebox{\columnwidth}{!}{
\begin{tabular}{@{}clrrrr@{}}
 &  &  & $c$1 & $c$2 & $c$3 \\ 
 \toprule 
 &
  & 
  & \multicolumn{1}{c}{\textbf{I} }
  & \multicolumn{2}{c}{\textbf{I+$R_\textrm{\ecqa}$} } \\
  \cmidrule(lr){4-4}\cmidrule(lr){5-6}
  & & \textbf{\tiny \%$R^\textrm{Tr}$}& & \textbf{w/o shuffle} & \textbf{shuffled} \\
  \midrule
$r1$ & I$\rightarrow$O & - & 57.00 & 53.32 & 54.95 \\ 
\midrule
$r2$  & \multirow{5}{*}{\rotatebox[origin=c]{90}{\tiny IR$^\textrm{Tr}_\textrm{\ecqa}\rightarrow$O}} & 5\% & 55.45 & \textbf{93.94} & \textbf{76.66} \\
$r3$  & & 10\% & 55.36 & 96.56 & 73.46 \\
$r4$  & & 20\% & 54.55 & 97.21 & 70.02 \\
$r5$  & & 30\% & 53.64 & 97.46 & 66.91 \\ 
$r6$  & & 100\% & 31.44 & 97.79 & 76.33 \\ 
\bottomrule
\end{tabular}
}
\caption{
The importance of shuffling the order of sentences in \ecqa rationales in training.
Without shuffling, the model relies on the spurious correlation due to sentence order, as compared to $r$7-11/$c4$ in  Tab.~\ref{tab:train-crowdR-comqa}.
Accuracies are averaged across 3 random seeds (s.d. as subscript) for \%R selection during training, as in Tab.~\ref{tab:train-crowdR-comqa}.
}
\label{tab:train-crowdR-ecqa-shuffle}
\end{table}

\section{Can Models Benefit from Crowdsourced Rationales?}
\label{sec:experiments}

In \S\ref{sec:human}, we found that crowdsourced rationales from carefully constructed corpora provide additional information to help humans better answer commonsense questions.
Now, we seek to answer if these rationales could also help in model learning, by providing an additional training signal to make better decisions, taking into account our findings from the detailed analysis in \S\ref{sec:human}.

\paragraph{Experimental Setup.} 
We use finetuned T5~\cite{t5} models throughout our work following prior efforts for analyzing~\cite{wiegreffe-etal-2021-measuring} and generating~\cite{Narang2020WT5TT,lakhotia-etal-2021-fid} free-text explanations. 
More specifically, we finetune three model classes based on the T5-base architecture: 
\begin{itemize}
    \item I$\rightarrow$O. Predict the label directly from the question and answer options.
    \item IR$\rightarrow$O. Predict the label from the question, answer options \textit{and} the rationale.
    \item I$\rightarrow$R. Predict the rationale from the question and answer options.
\end{itemize}
For the IR$\rightarrow$O model, we experiment with different variations based on the source, and the quantity of the rationales $R$, provided during training. 
Since most of our experiments deal with the first two model classes, we report accuracy of output label prediction.
See App.~\ref{app:t5} for details on our T5 model training and I/O formats.

We use rationales for the ComQA training instances to train two different sets of IR$\rightarrow$O models, for \cose and \ecqa respectively.
Under each set, we train five different models, randomly selecting different amounts (5\%, 10\%, 20\%, 30\% and the full 100\%) of \cose and shuffled-\ecqa rationales for training.\footnote{Some training instances receive both the I and $R$ and others receive just I, see Appendix~\ref{sec:format}.}
During training, we use varying amounts (5\%, 10\%, 20\%, 30\% and the full 100\%) of \cose and shuffled-\ecqa rationales, to study how the quantity of rationales affects the model performance.
During inference, we provide the IR$\rightarrow$O  T5 models with rationales under each of the four categories of \cose, as in Table~\ref{tab:cose-categorization}, as well as all combined together.
For \ecqa, we report performance for inference with and without shuffled rationales.
Finally, we study how rationales from one dataset transfer to the other.

\paragraph{Crowdsourced rationales boost model performance, ruling out leakage.} 
Comparing $c1$ in Table~\ref{tab:train-crowdR-comqa} with the columns $c2$-$c8$, we see that rationales help improve the model's ability to make the correct prediction, even when including only 5\% of the rationales during training.
However, instances that leak the answer make up a large portion of \cose. 
Indeed, when provided at test time, \emph{rationales which neither leak the correct answer nor provide additional background knowledge, cause the least improvement in model performance ($c7$).}
Further, with background knowledge, but no leakage, model performance can still be improved ($c5$); after adding 5\% of the training data, the model reaches 59.89\% accuracy with $\small{C_{\textrm{no-leak-bg}}}$ rationales, which yields 47.2\% improvement, compared to 40.68\% without rationales.\footnote{Unlike test rationales from other categories, the trends are not monotonic for $R_\textrm{no-leak-bg}$, most likely because this is the smallest (only 4\%) subset of the test set (Table~\ref{tab:cose-categorization}). 
\swabha{need another pair of eyes for this footnote.}
}
Overall, a close inspection of the rationales is necessary to understand when they can help the model decision for the right reasons (i.e. providing background information, not simply by leaking the answer). 
In other words, models can benefit from those crowdsourced rationales which provide utility for human interpretability as well!



\paragraph{Not all rationales are the same.}
We see benefits from increasing the amount of \ecqa rationales in the training data ($r7$-$r11$/ $c4$), even in a transfer setting ($r7$-$r11$/ $c2$).
However, this trend is weaker when training with \cose ($r2-r6$).
This highlights the importance of a rigorous procedure for crowdsourcing rationales \cite{aggarwal-etal-2021-explanations}.

\begin{table}[]
\small
\centering
\resizebox{\columnwidth}{!}{
\begin{tabular}{@{}llrll@{}}
\toprule
 & \multicolumn{1}{l}{} & \textbf{\%R$^\textrm{Tr}$} & \textbf{I} & \textbf{I+R$_\textrm{\quartz}$} \\ 
\midrule
r1 & \multicolumn{1}{l}{I$\rightarrow$O} & - & 70.88 & 38.27 \\ 
\midrule
r2 & \multicolumn{1}{l}{\multirow{5}{*}{\tiny \rotatebox[origin=c]{90}{IR$^\textrm{Tr}_\textrm{\quartz}\rightarrow$O}}} & 5\% & 66.20$_{1.33}$ & 67.86$_{1.18}$ \\
r3 & \multicolumn{1}{l}{} & 10\% & 67.81$_{1.15}$ & 70.58$_{1.25}$ \\
r4 & \multicolumn{1}{l}{} & 20\% & 67.99$_{0.54}$ & 69.73$_{0.97}$ \\
r5 & \multicolumn{1}{l}{} & 30\% & 67.13$_{0.69}$ & 71.51$_{0.16}$ \\
r6 & \multicolumn{1}{l}{} & 100\% & 64.67 & 81.51 \\ \bottomrule
\end{tabular}
}
\caption{
\quartz model accuracy with and without training with knowledge statements as rationales.
We report accuracies averaged across 3 random seeds (s.d. as subscript) for \%R selection during training, as in Table~\ref{tab:train-crowdR-comqa}.
}
\label{tab:train-crowdR-quartz}
\end{table}

\paragraph{Spurious correlations in rationales must be minimized.}
Recall from \S\ref{sec:categorizing} that \ecqa rationales tend to follow an ordering: sentences rationalizing the correct option precede those refuting the incorrect ones.
To validate the importance of shuffling sentences in \ecqa rationales, we present a baseline in Table~\ref{tab:train-crowdR-ecqa-shuffle} which considers unshuffled rationales during training, to be compared to training with shuffled rationales in Table~\ref{tab:train-crowdR-comqa}.
In the unshuffled case, training with only 5\% rationales improves the accuracy on unshuffled test rationales from 53.32\% to 93.94\% ($c2$, Tab.~\ref{tab:train-crowdR-ecqa-shuffle}).
However, when we test the same model using shuffled rationales, the accuracy improves from 54.95\% to 76.66\% ($c3$). 
This shows that the model might learn a spurious correlation between the rationale and correct answer, due to ordering.
We recommend shuffling \ecqa rationales before using them for model training.

\paragraph{Training with non-leaky rationales is beneficial.}
Despite taking care to prevent spurious correlations in \ecqa, there is still a chance that the models benefit from some amount of leakage of the correct answer, an uninteresting use of rationales to improve model performance.
To control for this, we consider the \quartz dataset, introduced in \S\ref{sec:tasks}, using knowledge statements as rationales, which are designed to contain no leakage, but provide the background information.
Using a similar setup to our ComQA experiments above, we finetune T5 models for both I$\rightarrow$O and IR$\rightarrow$O models on \quartz. 
Results in Table~\ref{tab:train-crowdR-quartz} show that the non-leaky \quartz rationales improve a model's ability to predict the correct answer, consistent with our findings in Table~\ref{tab:train-crowdR-comqa}.
These highlight the generalizability of our conclusions.

\begin{table}[t]
\centering
\small
\resizebox{\columnwidth}{!}{
\begin{tabular}{@{}llrrrr@{}}
\toprule
 & \textbf{} & \textbf{$R^\textrm{crowd}$} & \textbf{$R^\textrm{gen.}$} & \textbf{neither} & \textbf{both} \\ 
 \midrule
\multirow{2}{*}{\rotatebox{90}{\tiny \cose}} & Q1: has bg. knowl.? & 28.33\%  & 20.00\% & 34.17\%  & 17.50\%  \\ 
                       & Q2: leaks answer?             & 51.67\% & 40.83\% &  - & - \\ 
\midrule
\multirow{2}{*}{\rotatebox{90}{\tiny \ecqa}} & Q1: has bg. knowl.?  & 43.44\% & 22.50\% & 15.00\% & 19.17\% \\ 
                       & Q2: leaks answer?               & 89.17\% & 64.17\% & - & - \\ 

\bottomrule
\end{tabular}
}
\caption{
Comparative human studies for direct assessment of annotated vs. generated rationales in \cose and \ecqa, similar to Table~\ref{tab:human-direct}. 
Humans believe that generated rationales less frequently provide additional background knowledge than crowdsourced rationales. 
}
\label{tab:human-direct-generatedR}
\end{table}



\begin{table}[t]
\small
\centering
\resizebox{\columnwidth}{!}{
\begin{tabular}{@{}rrllll@{}}
\toprule
& & \multicolumn{2}{c}{testing with generated $R$} & \multicolumn{2}{c}{testing with crowd $R$} \\
\cmidrule(lr){3-4}\cmidrule(lr){5-6}
 & \textbf{\%$R^\textrm{gen.}_\textrm{\tiny Tr.}$} 
 & \textbf{$R^\textrm{gen.}_\textrm{\tiny \cose}$} 
 & \textbf{$R^\textrm{gen.}_\textrm{\tiny \ecqa}$}
 & \textbf{$R^\textrm{crowd}_\textrm{\tiny \cose}$}
 & \textbf{$R^\textrm{crowd}_\textrm{\tiny \ecqa}$}
 \\ \midrule
\multirow{5}{*}{\rotatebox[origin=c]{90}{\tiny IR$^\textrm{gen.}_\textrm{\cose}\rightarrow$O}} 
 & 5\% & 44.34$_{1.59}$ & 45.1$_{0.86}$ & 68.96$_{2.11}$ & 58.83$_{1.27}$\\
 & 10\% & 44.94$_{0.59}$ & 42.89$_{0.46}$ & 75.98$_{0.72}$ & 60.53$_{0.46}$\\
 & 20\% & 44.34$_{0.71}$ & 41.17$_{0.74}$ & 77.45$_{0.54}$ & 62.38$_{0.27}$\\
 & 30\% & 44.91$_{0.43}$ & 39.83$_{0.56}$ & 77.07$_{0.70}$ & 64.65$_{0.95}$\\
 & 100\% & \textbf{43.90} & 35.71 & \textbf{76.74} & 60.11\\ \midrule
\multirow{5}{*}{\rotatebox[origin=c]{90}{\tiny IR$^\textrm{gen.}_\textrm{\ecqa}\rightarrow$O}} 
 & 5\% & 46.33$_{0.54}$ & 44.64$_{1.03}$  & 58.97$_{1.07}$ & 79.31$_{1.17}$\\
 & 10\% & 45.10$_{0.34}$ & 44.96$_{0.30}$ & 60.41$_{0.47}$ & 87.22$_{0.40}$\\
 & 20\% & 46.98$_{0.83}$ & 45.67$_{0.37}$ & 62.00$_{0.35}$ & 89.68$_{0.47}$\\
 & 30\% & 45.81$_{0.60}$ & 45.51$_{0.40}$ & 64.51$_{1.03}$ & 91.51$_{0.80}$\\
 & 100\% & 43.16 & \textbf{44.64} & 64.86 & \textbf{93.37}\\ \bottomrule
\end{tabular}
}
\caption{
Performance of IR$\rightarrow$O models, trained with different amounts (\%$R^\textrm{gen.}_\textrm{\tiny Tr.}$) of generated rationales.
The top block indicates training with rationales generated from a \cose-trained I$\rightarrow$R model, and the bottom block, an \ecqa-trained model.
The columns indicate rationales provided at test time to the IR$\rightarrow$O models.
We report accuracies averaged across 3 random seeds (s.d. as subscript) for \%$R^\textrm{gen.}_\textrm{\tiny Tr.}$ selection, as in Table~\ref{tab:train-crowdR-comqa}.
}
\label{tab:train-generatedR-comqa}
\end{table}

\section{Can Models Benefit from Generated Rationales?}
\label{sec:generated}
So far, we have focused on crowdsourced rationales, the primary reason behind collecting which is to train models that generate them automatically, as seen in recent work \cite{Narang2020WT5TT, paranjape-etal-2021-prompting}. 
Hence, we now ask: (1) analogous to crowdsourced rationales (\S\ref{sec:human}), can \textit{generated} rationales provide the additional background information necessary for human interpretability, and (2) can \textit{generated} rationales provide additional training signals to improve model performance, similar to \S\ref{sec:experiments}?

Using crowdsourced \cose and \ecqa rationales as supervision, we train two different T5-base I$\rightarrow$R models, following \S\ref{sec:tasks}.
We generate rationales on all ComQA instances (train as well as test) using these two models.\footnote{
A qualitative analysis of annotated and generated rationales for both \cose and \ecqa can be found in Appendix~\ref{app:generation-example}.
} 

\paragraph{Human Perception of Generated Rationales.} 
\label{app:subjective}

We repeat our comparative human study via direct assessment (on the same 120 test ComQA instances from \S\ref{sec:human-direct}) on generated vs. annotated (1) \ecqa and (2) \cose rationales (see Fig.~\ref{fig:human-direct} in Appendix~\ref{app:anno-human-study}).
Table~\ref{tab:human-direct-generatedR} shows that humans believe generated rationales provide background knowledge less often than human-annotated rationales.\footnote{
For generated rationales, attributes such as plausibility and faithfulness \cite{jacovi-goldberg-2021-aligning} also matter. 
However, we focus on the same attributes as we use to evaluate crowdsourced rationales (\S\ref{sec:human}), for a contrastive comparison.}
Generated \ecqa rationales have more background knowledge than generated \cose rationales, but do leak the correct answer more often as well, reflecting the \ecqa training data. 
These results are consistent with our findings in \S\ref{sec:human}.

\vspace{1mm}
\paragraph{Training with Generated Rationales.}
Next, we use generated rationales for the ComQA training instances to train two different sets of IR$\rightarrow$O models, for \cose and \ecqa respectively.
Under each set, we train five different models, randomly selecting different amounts of generated rationales for training.
At inference time, we provide the IR$\rightarrow$O models, either generated rationales for each ComQA test instance, following the setting from \citet{wiegreffe-etal-2021-measuring}, or crowdsourced rationales for the same set.

Table~\ref{tab:train-generatedR-comqa} shows the results. 
When testing with generated rationales, we see a reduction in the model's predictive capability compared to testing with crowdsourced rationales.\footnote{
See App.~\S\ref{sec:simulatability} for an evaluation of these rationales.
}
Moreover, training with \ecqa-generated rationales seems more beneficial than training with \cose-generated rationales.
Training with larger quantities of generated rationales keeps boosting the model performance to an extent (from 5\% to 30\%), consistent to our findings with crowdsourced rationales. 
However, when each and every instance is paired with its generated rationale during training (rows with 100\%), the performance drops when testing with generated rationales as well.
We suspect this might be attributed to the generated rationales introducing too much noise, and not providing the model a clear signal. 
Regardless, under neither setting, do these models do as well as the models trained with crowdsourced rationales, as shown in Table~\ref{tab:train-crowdR-comqa}.


\section{Related Work}
\label{sec:related}

Rationales serve interpretability in that they can reveal the ``reasoning'' behind model decisions, and can be roughly categorized into two broad categories: extractive and free-form rationales. 
Extractive rationales provide supportive evidence in a grounded context, such as textual highlights within an input document \cite{lei-etal-2016-rationalizing,deyoung-etal-2020-eraser}, sufficient to make a prediction on its own without relying on the rest of the input. 
Analogous to our work, there has been a line of work that studies the value of extractive rationales as additional training signals to improve model performance \cite{huang-etal-2021-exploring,Carton2021WhatTL} or human interpretability \cite{strout-etal-2019-human}.
We use free-text rationales, on the other hand, which employ free-form natural language to fill in the commonsense reasoning or knowledge gaps.
Such rationales have been used for language and vision tasks \cite{hendricks2016generating,kim2018textual} but have far less adoption in NLP \cite{Wiegreffe2021TeachMT}.
Concurrent to our work, \citet{hase-bansal-2022-models} provide a formal framework for free-text explanation utility for models, and a synthetic dataset for the same.
In addition to the datasets used in this work, e-SNLI~\cite{Camburu2018eSNLINL} provides free-text rationales for the natural language inference task.


Rationale generation models can be roughly categorized into supervised \cite{lakhotia-etal-2021-fid, Narang2020WT5TT,kumar2020nile, rajani-etal-2019-explain, Zhao2021lirex} and unsupervised \cite{glockner-etal-2020-think,brahman2021learning}. 
For supervised models, \citet{lakhotia-etal-2021-fid} and \citet{Narang2020WT5TT} finetune T5 to generate extractive and free-form rationales separately. 
For unsupervised models, \citet{glockner-etal-2020-think} propose a differential training framework to create models that output faithful rationales without supervision. 
Instead of directly generating rationales, \citet{paranjape-etal-2021-prompting} use T5 to complete contrastive explanation prompts which explicitly contrast different possible answers \cite{jacovi-etal-2021-contrastive}. 
Recent work has relied on few-shot prompting to generate explanations \cite{wiegreffe2022reframing,marasovi2022fewshot}.
Our work follows a supervised approach to generate rationales, and further uses both generated and crowdsourced rationales as training signals to produce output labels.



\section{Conclusion}
\label{sec:conclusion}

We investigated the utility of free-form rationales from both a human and a modeling perspective. 
Centering our analysis on commonsense QA datasets,
we find that humans perceive rationales with more background knowledge as more useful than those which simply combine the question and the answer. 
We provided a detailed qualitative analysis of \cose and \ecqa rationales, and found that even small amounts of carefully written rationales are helpful as additional training signals for task models.
Our work highlights the importance of inspecting the quality of crowdsourced rationales before using them for additional supervision.
We also found that generated rationales are not as useful as crowdsourced rationales for human interpretability or for model supervision.
Our investigations shed light on fundamental assumptions about human interpretability in collecting and generating rationales, and call for further deeper investigation into the utility of free-form rationales.

\swabha{TODO: revisit our ARR reviews.}

\section*{Ethical Consideration}
\label{sec:ethical}
During our manual annotation process, we provide timely warning of potential adult topics and ask workers to return the job if they are under age. 
Our human studies are labeled as exempt from review by the IRB at the authors' institute.

For modeling, we use T5 throughout our work, which also involves generating rationales. 
It is well-known that such pretrained language models---trained on massive online texts---capture the biases reflected in the training data \cite{bender2021parrots}.
Therefore, the generation models can be used for malicious purposes and generate rationales that contain toxic content that target at certain groups.
We do not have a filtering mechanism that checks the toxicity, bias, or offensiveness of our training data, or that of our generated explanations.
Hence, we recommend practitioners interested in using and replicating this work to carefully check the generated content before deployment in any real world application. 

The datasets used in our work are all public. 
These do not contain any explicit detail that leaks information about a user's name, health, negative financial status, racial or ethnic origin, religious or philosophical affiliation or beliefs.

\section*{Limitations}
\label{sec:limitations}
This work is subject to several limitations. 
First, our human studies are pilot studies, where we annotated only 100 instances to understand how much rationales can aid the human interpretability. 
Although the trend we observed is intuitive and consistent, more data for the human study might improve the quality of the findings. 
Secondly, we use vanilla T5 models for rationale generation for a fair comparison with previous work. 
However, there could potentially be rationales of higher quality via more sophisticated and powerful language models, which are beyond the scope of our exploration.  
Last, this work focuses on question-answering for commonsense knowledge. 
It still remains unexplored if our conclusions transfer beyond this task.

There are many other ways to evaluate the utility of free-form rationales. Our work focuses on the specific aspect of if a rationale provides additional background information that can help answer the question. Please note that leakage, which is also studied in our work, does not reduce the utility of a rationale for human interpretability. It is natural for an explanation to explicitly lead to the correct answer. However, the specific reason why we include the study of leakage in \S\ref{sec:human} is that we wanted to bring to the annotator’s attention (implicitly) that a rationale might look good simply because it mentions the correct answer even though it might not contain the reasoning for it. We refer interested workers to \citet{jacovi-goldberg-2020-towards} for discussions about the criteria that constitutes a high-quality interpretation.  Future works can explore other aspects that can contribute to or jeopardize the utility of rationales for both humans and models, including but not limited to the factuality, completeness and presence of unnecessary information in rationales.  

\section*{Acknowledgments}
The authors thank anonymous reviewers and the meta reviewer for their constructive feedback and suggestions that helped improve the draft. This material is based upon work supported by the Defense Advanced Research Projects Agency (DARPA) under Agreement Nos. HR00112290025 and HR00112290056.

\bibliography{anthology,custom}

\begin{thebibliography}{49}
\expandafter\ifx\csname natexlab\endcsname\relax\def\natexlab#1{#1}\fi

\bibitem[{Aggarwal et~al.(2021)Aggarwal, Mandowara, Agrawal, Khandelwal,
  Singla, and Garg}]{aggarwal-etal-2021-explanations}
Shourya Aggarwal, Divyanshu Mandowara, Vishwajeet Agrawal, Dinesh Khandelwal,
  Parag Singla, and Dinesh Garg. 2021.
\newblock \href {https://doi.org/10.18653/v1/2021.acl-long.238} {{E}xplanations
  for {C}ommonsense{QA}: {N}ew {D}ataset and {M}odels}.
\newblock In \emph{Proceedings of the 59th Annual Meeting of the Association
  for Computational Linguistics and the 11th International Joint Conference on
  Natural Language Processing (Volume 1: Long Papers)}, pages 3050--3065,
  Online. Association for Computational Linguistics.

\bibitem[{Alvarez-Melis and Jaakkola(2018)}]{AlvarezMelis2018TowardsRI}
David Alvarez-Melis and T.~Jaakkola. 2018.
\newblock \href
  {https://proceedings.neurips.cc/paper/2018/hash/3e9f0fc9b2f89e043bc6233994dfcf76-Abstract.html}
  {Towards robust interpretability with self-explaining neural networks}.
\newblock In \emph{NeurIPS}.

\bibitem[{Bender et~al.(2021)Bender, Gebru, McMillan-Major, and
  Shmitchell}]{bender2021parrots}
Emily~M. Bender, Timnit Gebru, Angelina McMillan-Major, and Shmargaret
  Shmitchell. 2021.
\newblock \href {https://doi.org/10.1145/3442188.3445922} {On the dangers of
  stochastic parrots: Can language models be too big?}
\newblock In \emph{Proceedings of the 2021 ACM Conference on Fairness,
  Accountability, and Transparency}, FAccT '21, page 610–623, New York, NY,
  USA. Association for Computing Machinery.

\bibitem[{Brahman et~al.(2021)Brahman, Shwartz, Rudinger, and
  Choi}]{brahman2021learning}
Faeze Brahman, Vered Shwartz, Rachel Rudinger, and Yejin Choi. 2021.
\newblock \href {https://arxiv.org/abs/2012.08012} {Learning to rationalize for
  nonmonotonic reasoning with distant supervision}.
\newblock \emph{Proceedings of the AAAI Conference on Artificial Intelligence},
  35(14):12592--12601.

\bibitem[{Camburu et~al.(2018)Camburu, Rockt{\"a}schel, Lukasiewicz, and
  Blunsom}]{Camburu2018eSNLINL}
Oana-Maria Camburu, Tim Rockt{\"a}schel, Thomas Lukasiewicz, and Phil Blunsom.
  2018.
\newblock e-snli: Natural language inference with natural language
  explanations.
\newblock In \emph{NeurIPS}.

\bibitem[{Carton et~al.(2021)Carton, Kanoria, and Tan}]{Carton2021WhatTL}
Samuel Carton, Surya Kanoria, and Chenhao Tan. 2021.
\newblock What to learn, and how: Toward effective learning from rationales.
\newblock \emph{ArXiv}, abs/2112.00071.

\bibitem[{Clinciu et~al.(2021)Clinciu, Eshghi, and
  Hastie}]{clinciu-etal-2021-study}
Miruna-Adriana Clinciu, Arash Eshghi, and Helen Hastie. 2021.
\newblock \href {https://doi.org/10.18653/v1/2021.eacl-main.202} {A study of
  automatic metrics for the evaluation of natural language explanations}.
\newblock In \emph{Proceedings of the 16th Conference of the European Chapter
  of the Association for Computational Linguistics: Main Volume}, pages
  2376--2387, Online. Association for Computational Linguistics.

\bibitem[{DeYoung et~al.(2020)DeYoung, Jain, Rajani, Lehman, Xiong, Socher, and
  Wallace}]{deyoung-etal-2020-eraser}
Jay DeYoung, Sarthak Jain, Nazneen~Fatema Rajani, Eric Lehman, Caiming Xiong,
  Richard Socher, and Byron~C. Wallace. 2020.
\newblock \href {https://doi.org/10.18653/v1/2020.acl-main.408} {{ERASER}: {A}
  benchmark to evaluate rationalized {NLP} models}.
\newblock In \emph{Proceedings of the 58th Annual Meeting of the Association
  for Computational Linguistics}, pages 4443--4458, Online. Association for
  Computational Linguistics.

\bibitem[{Fleiss and Cohen(1973)}]{Fleiss1973TheEO}
Joseph~L. Fleiss and Jacob Cohen. 1973.
\newblock The equivalence of weighted kappa and the intraclass correlation
  coefficient as measures of reliability.
\newblock \emph{Educational and Psychological Measurement}, 33:613 -- 619.

\bibitem[{Geirhos et~al.(2020)Geirhos, Jacobsen, Michaelis, Zemel, Brendel,
  Bethge, and Wichmann}]{geirhos2020shortcut}
Robert Geirhos, Jörn-Henrik Jacobsen, Claudio Michaelis, Richard Zemel,
  Wieland Brendel, Matthias Bethge, and Felix~A. Wichmann. 2020.
\newblock \href {https://doi.org/10.1038/s42256-020-00257-z} {Shortcut learning
  in deep neural networks}.
\newblock \emph{Nature Machine Intelligence}, 2(11):665–673.

\bibitem[{Glockner et~al.(2020)Glockner, Habernal, and
  Gurevych}]{glockner-etal-2020-think}
Max Glockner, Ivan Habernal, and Iryna Gurevych. 2020.
\newblock \href {https://doi.org/10.18653/v1/2020.findings-emnlp.97} {Why do
  you think that? exploring faithful sentence-level rationales without
  supervision}.
\newblock In \emph{Findings of the Association for Computational Linguistics:
  EMNLP 2020}, pages 1080--1095, Online. Association for Computational
  Linguistics.

\bibitem[{Hase and Bansal(2020)}]{hase-bansal-2020-evaluating}
Peter Hase and Mohit Bansal. 2020.
\newblock \href {https://doi.org/10.18653/v1/2020.acl-main.491} {Evaluating
  explainable {AI}: Which algorithmic explanations help users predict model
  behavior?}
\newblock In \emph{Proceedings of the 58th Annual Meeting of the Association
  for Computational Linguistics}, pages 5540--5552, Online. Association for
  Computational Linguistics.

\bibitem[{Hase and Bansal(2022)}]{hase-bansal-2022-models}
Peter Hase and Mohit Bansal. 2022.
\newblock \href {https://doi.org/10.18653/v1/2022.lnls-1.4} {When can models
  learn from explanations? a formal framework for understanding the roles of
  explanation data}.
\newblock In \emph{Proceedings of the First Workshop on Learning with Natural
  Language Supervision}, pages 29--39, Dublin, Ireland. Association for
  Computational Linguistics.

\bibitem[{Hase et~al.(2020)Hase, Zhang, Xie, and
  Bansal}]{hase-etal-2020-leakage}
Peter Hase, Shiyue Zhang, Harry Xie, and Mohit Bansal. 2020.
\newblock \href {https://doi.org/10.18653/v1/2020.findings-emnlp.390}
  {Leakage-adjusted simulatability: Can models generate non-trivial
  explanations of their behavior in natural language?}
\newblock In \emph{Findings of the Association for Computational Linguistics:
  EMNLP 2020}, pages 4351--4367, Online. Association for Computational
  Linguistics.

\bibitem[{Hendricks et~al.(2016)Hendricks, Akata, Rohrbach, Donahue, Schiele,
  and Darrell}]{hendricks2016generating}
Lisa~Anne Hendricks, Zeynep Akata, Marcus Rohrbach, Jeff Donahue, Bernt
  Schiele, and Trevor Darrell. 2016.
\newblock \href {https://doi.org/10.1007/978-3-319-46493-0_1} {Generating
  visual explanations}.
\newblock \emph{Lecture Notes in Computer Science}, page 3–19.

\bibitem[{Hohman et~al.(2020)Hohman, Park, Robinson, and
  Chau}]{Hohman2020SummitSD}
Fred Hohman, Haekyu Park, Caleb Robinson, and Duen~Horng Chau. 2020.
\newblock Summit: Scaling deep learning interpretability by visualizing
  activation and attribution summarizations.
\newblock \emph{IEEE Transactions on Visualization and Computer Graphics},
  26:1096--1106.

\bibitem[{Huang et~al.(2021)Huang, Zhu, Feng, and
  Zhao}]{huang-etal-2021-exploring}
Quzhe Huang, Shengqi Zhu, Yansong Feng, and Dongyan Zhao. 2021.
\newblock \href {https://doi.org/10.18653/v1/2021.acl-long.433} {Exploring
  distantly-labeled rationales in neural network models}.
\newblock In \emph{Proceedings of the 59th Annual Meeting of the Association
  for Computational Linguistics and the 11th International Joint Conference on
  Natural Language Processing (Volume 1: Long Papers)}, pages 5571--5582,
  Online. Association for Computational Linguistics.

\bibitem[{Jacovi and Goldberg(2020)}]{jacovi-goldberg-2020-towards}
Alon Jacovi and Yoav Goldberg. 2020.
\newblock \href {https://doi.org/10.18653/v1/2020.acl-main.386} {Towards
  faithfully interpretable {NLP} systems: How should we define and evaluate
  faithfulness?}
\newblock In \emph{Proceedings of the 58th Annual Meeting of the Association
  for Computational Linguistics}, pages 4198--4205, Online. Association for
  Computational Linguistics.

\bibitem[{Jacovi and Goldberg(2021)}]{jacovi-goldberg-2021-aligning}
Alon Jacovi and Yoav Goldberg. 2021.
\newblock \href {https://doi.org/10.1162/tacl_a_00367} {Aligning faithful
  interpretations with their social attribution}.
\newblock \emph{Transactions of the Association for Computational Linguistics},
  9:294--310.

\bibitem[{Jacovi et~al.(2021)Jacovi, Swayamdipta, Ravfogel, Elazar, Choi, and
  Goldberg}]{jacovi-etal-2021-contrastive}
Alon Jacovi, Swabha Swayamdipta, Shauli Ravfogel, Yanai Elazar, Yejin Choi, and
  Yoav Goldberg. 2021.
\newblock \href {https://aclanthology.org/2021.emnlp-main.120} {Contrastive
  explanations for model interpretability}.
\newblock In \emph{Proceedings of the 2021 Conference on Empirical Methods in
  Natural Language Processing}, pages 1597--1611, Online and Punta Cana,
  Dominican Republic. Association for Computational Linguistics.

\bibitem[{Kayser et~al.(2021)Kayser, Camburu, Salewski, Emde, Do, Akata, and
  Lukasiewicz}]{Kayser2021eViLAD}
Maxime Kayser, Oana{-}Maria Camburu, Leonard Salewski, Cornelius Emde, Virginie
  Do, Zeynep Akata, and Thomas Lukasiewicz. 2021.
\newblock \href {https://doi.org/10.1109/ICCV48922.2021.00128} {e-vil: {A}
  dataset and benchmark for natural language explanations in vision-language
  tasks}.
\newblock In \emph{2021 {IEEE/CVF} International Conference on Computer Vision,
  {ICCV} 2021, Montreal, QC, Canada, October 10-17, 2021}, pages 1224--1234.
  {IEEE}.

\bibitem[{Kim(2015)}]{kim2015interactive}
Been Kim. 2015.
\newblock \href {https://dspace.mit.edu/handle/1721.1/98680} {\emph{Interactive
  and interpretable machine learning models for human machine collaboration}}.
\newblock Ph.D. thesis, Massachusetts Institute of Technology.

\bibitem[{Kim et~al.(2018)Kim, Rohrbach, Darrell, Canny, and
  Akata}]{kim2018textual}
Jinkyu Kim, Anna Rohrbach, Trevor Darrell, John Canny, and Zeynep Akata. 2018.
\newblock \href {https://doi.org/10.1007/978-3-030-01216-8_35} {Textual
  explanations for self-driving vehicles}.
\newblock \emph{Lecture Notes in Computer Science}, page 577–593.

\bibitem[{Kumar and Talukdar(2020)}]{kumar2020nile}
Sawan Kumar and Partha Talukdar. 2020.
\newblock \href {https://doi.org/10.18653/v1/2020.acl-main.771} {{NILE} :
  Natural language inference with faithful natural language explanations}.
\newblock In \emph{Proceedings of the 58th Annual Meeting of the Association
  for Computational Linguistics}, pages 8730--8742, Online. Association for
  Computational Linguistics.

\bibitem[{Lakhotia et~al.(2021)Lakhotia, Paranjape, Ghoshal, Yih, Mehdad, and
  Iyer}]{lakhotia-etal-2021-fid}
Kushal Lakhotia, Bhargavi Paranjape, Asish Ghoshal, Scott Yih, Yashar Mehdad,
  and Srini Iyer. 2021.
\newblock \href {https://aclanthology.org/2021.emnlp-main.301} {{F}i{D}-ex:
  Improving sequence-to-sequence models for extractive rationale generation}.
\newblock In \emph{Proceedings of the 2021 Conference on Empirical Methods in
  Natural Language Processing}, pages 3712--3727, Online and Punta Cana,
  Dominican Republic. Association for Computational Linguistics.

\bibitem[{Lei et~al.(2016)Lei, Barzilay, and
  Jaakkola}]{lei-etal-2016-rationalizing}
Tao Lei, Regina Barzilay, and Tommi Jaakkola. 2016.
\newblock \href {https://doi.org/10.18653/v1/D16-1011} {Rationalizing neural
  predictions}.
\newblock In \emph{Proceedings of the 2016 Conference on Empirical Methods in
  Natural Language Processing}, pages 107--117, Austin, Texas. Association for
  Computational Linguistics.

\bibitem[{Lipton(2018)}]{lipton2018mythos}
Zachary~C Lipton. 2018.
\newblock \href {https://queue.acm.org/detail.cfm?id=3241340} {The mythos of
  model interpretability: In machine learning, the concept of interpretability
  is both important and slippery.}
\newblock \emph{Queue}, 16(3):31--57.

\bibitem[{Loshchilov and Hutter(2019)}]{adamw}
Ilya Loshchilov and Frank Hutter. 2019.
\newblock \href {https://openreview.net/forum?id=Bkg6RiCqY7} {Decoupled weight
  decay regularization}.
\newblock In \emph{7th International Conference on Learning Representations,
  {ICLR} 2019, New Orleans, LA, USA, May 6-9, 2019}. OpenReview.net.

\bibitem[{Majumder et~al.(2021)Majumder, Camburu, Lukasiewicz, and
  McAuley}]{Majumder2021RationaleInspiredNL}
Bodhisattwa~Prasad Majumder, Oana-Maria Camburu, Thomas Lukasiewicz, and Julian
  McAuley. 2021.
\newblock Rationale-inspired natural language explanations with commonsense.
\newblock \emph{ArXiv}, abs/2106.13876.

\bibitem[{Marasović et~al.(2022)Marasović, Beltagy, Downey, and
  Peters}]{marasovi2022fewshot}
Ana Marasović, Iz~Beltagy, Doug Downey, and Matthew~E. Peters. 2022.
\newblock \href {https://arxiv.org/abs/2111.08284} {Few-shot
  self-rationalization with natural language prompts}.
\newblock In \emph{Findings of NAACL}.

\bibitem[{Narang et~al.(2020)Narang, Raffel, Lee, Roberts, Fiedel, and
  Malkan}]{Narang2020WT5TT}
Sharan Narang, Colin Raffel, Katherine Lee, Adam Roberts, Noah Fiedel, and
  Karishma Malkan. 2020.
\newblock \href {https://arxiv.org/abs/2004.14546} {Wt5?! training text-to-text
  models to explain their predictions}.
\newblock \emph{ArXiv}, abs/2004.14546.

\bibitem[{Papineni et~al.(2002)Papineni, Roukos, Ward, and
  Zhu}]{papineni-etal-2002-bleu}
Kishore Papineni, Salim Roukos, Todd Ward, and Wei-Jing Zhu. 2002.
\newblock \href {https://doi.org/10.3115/1073083.1073135} {{B}leu: a method for
  automatic evaluation of machine translation}.
\newblock In \emph{Proceedings of the 40th Annual Meeting of the Association
  for Computational Linguistics}, pages 311--318, Philadelphia, Pennsylvania,
  USA. Association for Computational Linguistics.

\bibitem[{Paranjape et~al.(2021)Paranjape, Michael, Ghazvininejad, Hajishirzi,
  and Zettlemoyer}]{paranjape-etal-2021-prompting}
Bhargavi Paranjape, Julian Michael, Marjan Ghazvininejad, Hannaneh Hajishirzi,
  and Luke Zettlemoyer. 2021.
\newblock \href {https://doi.org/10.18653/v1/2021.findings-acl.366} {Prompting
  contrastive explanations for commonsense reasoning tasks}.
\newblock In \emph{Findings of the Association for Computational Linguistics:
  ACL-IJCNLP 2021}, pages 4179--4192, Online. Association for Computational
  Linguistics.

\bibitem[{Poursabzi-Sangdeh et~al.(2021)Poursabzi-Sangdeh, Goldstein, Hofman,
  Vaughan, and Wallach}]{PoursabziSangdeh2021ManipulatingAM}
Forough Poursabzi-Sangdeh, Daniel~G. Goldstein, Jake~M. Hofman,
  Jennifer~Wortman Vaughan, and Hanna~M. Wallach. 2021.
\newblock \href {https://arxiv.org/abs/1802.07810} {Manipulating and measuring
  model interpretability}.
\newblock \emph{Proceedings of the 2021 CHI Conference on Human Factors in
  Computing Systems}.

\bibitem[{Raffel et~al.(2020)Raffel, Shazeer, Roberts, Lee, Narang, Matena,
  Zhou, Li, and Liu}]{t5}
Colin Raffel, Noam Shazeer, Adam Roberts, Katherine Lee, Sharan Narang, Michael
  Matena, Yanqi Zhou, Wei Li, and Peter~J. Liu. 2020.
\newblock \href {http://jmlr.org/papers/v21/20-074.html} {Exploring the limits
  of transfer learning with a unified text-to-text transformer}.
\newblock \emph{Journal of Machine Learning Research}, 21(140):1--67.

\bibitem[{Rajagopal et~al.(2021)Rajagopal, Balachandran, Hovy, and
  Tsvetkov}]{rajagopal-etal-2021-selfexplain}
Dheeraj Rajagopal, Vidhisha Balachandran, Eduard~H Hovy, and Yulia Tsvetkov.
  2021.
\newblock \href {https://aclanthology.org/2021.emnlp-main.64} {{SELFEXPLAIN}: A
  self-explaining architecture for neural text classifiers}.
\newblock In \emph{Proceedings of the 2021 Conference on Empirical Methods in
  Natural Language Processing}, pages 836--850, Online and Punta Cana,
  Dominican Republic. Association for Computational Linguistics.

\bibitem[{Rajani et~al.(2019)Rajani, McCann, Xiong, and
  Socher}]{rajani-etal-2019-explain}
Nazneen~Fatema Rajani, Bryan McCann, Caiming Xiong, and Richard Socher. 2019.
\newblock \href {https://doi.org/10.18653/v1/P19-1487} {Explain yourself!
  leveraging language models for commonsense reasoning}.
\newblock In \emph{Proceedings of the 57th Annual Meeting of the Association
  for Computational Linguistics}, pages 4932--4942, Florence, Italy.
  Association for Computational Linguistics.

\bibitem[{Sap et~al.(2020)Sap, Gabriel, Qin, Jurafsky, Smith, and
  Choi}]{sap-etal-2020-social}
Maarten Sap, Saadia Gabriel, Lianhui Qin, Dan Jurafsky, Noah~A. Smith, and
  Yejin Choi. 2020.
\newblock \href {https://doi.org/10.18653/v1/2020.acl-main.486} {Social bias
  frames: Reasoning about social and power implications of language}.
\newblock In \emph{Proceedings of the 58th Annual Meeting of the Association
  for Computational Linguistics}, pages 5477--5490, Online. Association for
  Computational Linguistics.

\bibitem[{Strout et~al.(2019)Strout, Zhang, and
  Mooney}]{strout-etal-2019-human}
Julia Strout, Ye~Zhang, and Raymond Mooney. 2019.
\newblock \href {https://doi.org/10.18653/v1/W19-4807} {Do human rationales
  improve machine explanations?}
\newblock In \emph{Proceedings of the 2019 ACL Workshop BlackboxNLP: Analyzing
  and Interpreting Neural Networks for NLP}, pages 56--62, Florence, Italy.
  Association for Computational Linguistics.

\bibitem[{Sun et~al.(2021)Sun, Ma, and Peng}]{sun-etal-2021-aesop}
Jiao Sun, Xuezhe Ma, and Nanyun Peng. 2021.
\newblock \href {https://aclanthology.org/2021.emnlp-main.420} {{AESOP}:
  Paraphrase generation with adaptive syntactic control}.
\newblock In \emph{Proceedings of the 2021 Conference on Empirical Methods in
  Natural Language Processing}, pages 5176--5189, Online and Punta Cana,
  Dominican Republic. Association for Computational Linguistics.

\bibitem[{Tafjord et~al.(2019)Tafjord, Gardner, Lin, and
  Clark}]{tafjord-etal-2019-quartz}
Oyvind Tafjord, Matt Gardner, Kevin Lin, and Peter Clark. 2019.
\newblock \href {https://doi.org/10.18653/v1/D19-1608} {{Q}ua{RT}z: An
  open-domain dataset of qualitative relationship questions}.
\newblock In \emph{Proceedings of the 2019 Conference on Empirical Methods in
  Natural Language Processing and the 9th International Joint Conference on
  Natural Language Processing (EMNLP-IJCNLP)}, pages 5941--5946, Hong Kong,
  China. Association for Computational Linguistics.

\bibitem[{Tan(2022)}]{tan2021diversity}
Chenhao Tan. 2022.
\newblock \href {https://doi.org/10.18653/v1/2022.naacl-main.158} {On the
  diversity and limits of human explanations}.
\newblock In \emph{Proceedings of the 2022 Conference of the North American
  Chapter of the Association for Computational Linguistics: Human Language
  Technologies}, pages 2173--2188, Seattle, United States. Association for
  Computational Linguistics.

\bibitem[{Trajanovski et~al.(2021)Trajanovski, Atalla, Kim, Agarwal, Shokouhi,
  and Quirk}]{trajanovski-etal-2021-text}
Stojan Trajanovski, Chad Atalla, Kunho Kim, Vipul Agarwal, Milad Shokouhi, and
  Chris Quirk. 2021.
\newblock \href {https://doi.org/10.18653/v1/2021.naacl-industry.1} {When does
  text prediction benefit from additional context? an exploration of contextual
  signals for chat and email messages}.
\newblock In \emph{Proceedings of the 2021 Conference of the North American
  Chapter of the Association for Computational Linguistics: Human Language
  Technologies: Industry Papers}, pages 1--9, Online. Association for
  Computational Linguistics.

\bibitem[{Wiegreffe et~al.(2022)Wiegreffe, Hessel, Swayamdipta, Riedl, and
  Choi}]{wiegreffe2022reframing}
Sarah Wiegreffe, Jack Hessel, Swabha Swayamdipta, Mark Riedl, and Yejin Choi.
  2022.
\newblock \href {https://doi.org/10.18653/v1/2022.naacl-main.47} {Reframing
  human-{AI} collaboration for generating free-text explanations}.
\newblock In \emph{Proceedings of the 2022 Conference of the North American
  Chapter of the Association for Computational Linguistics: Human Language
  Technologies}, pages 632--658, Seattle, United States. Association for
  Computational Linguistics.

\bibitem[{Wiegreffe and Marasovic(2021)}]{Wiegreffe2021TeachMT}
Sarah Wiegreffe and Ana Marasovic. 2021.
\newblock \href {https://openreview.net/forum?id=ogNcxJn32BZ} {Teach me to
  explain: A review of datasets for explainable natural language processing}.
\newblock In \emph{Thirty-fifth Conference on Neural Information Processing
  Systems Datasets and Benchmarks Track (Round 1)}.

\bibitem[{Wiegreffe et~al.(2021)Wiegreffe, Marasovi{\'c}, and
  Smith}]{wiegreffe-etal-2021-measuring}
Sarah Wiegreffe, Ana Marasovi{\'c}, and Noah~A. Smith. 2021.
\newblock \href {https://aclanthology.org/2021.emnlp-main.804} {{M}easuring
  association between labels and free-text rationales}.
\newblock In \emph{Proceedings of the 2021 Conference on Empirical Methods in
  Natural Language Processing}, pages 10266--10284, Online and Punta Cana,
  Dominican Republic. Association for Computational Linguistics.

\bibitem[{Wolf et~al.(2020)Wolf, Debut, Sanh, Chaumond, Delangue, Moi, Cistac,
  Rault, Louf, Funtowicz, Davison, Shleifer, von Platen, Ma, Jernite, Plu, Xu,
  Le~Scao, Gugger, Drame, Lhoest, and Rush}]{wolf-etal-2020-transformers}
Thomas Wolf, Lysandre Debut, Victor Sanh, Julien Chaumond, Clement Delangue,
  Anthony Moi, Pierric Cistac, Tim Rault, Remi Louf, Morgan Funtowicz, Joe
  Davison, Sam Shleifer, Patrick von Platen, Clara Ma, Yacine Jernite, Julien
  Plu, Canwen Xu, Teven Le~Scao, Sylvain Gugger, Mariama Drame, Quentin Lhoest,
  and Alexander Rush. 2020.
\newblock \href {https://doi.org/10.18653/v1/2020.emnlp-demos.6} {Transformers:
  State-of-the-art natural language processing}.
\newblock In \emph{Proceedings of the 2020 Conference on Empirical Methods in
  Natural Language Processing: System Demonstrations}, pages 38--45, Online.
  Association for Computational Linguistics.

\bibitem[{Ye et~al.(2019)Ye, Chen, Wang, and Ling}]{Ye2019AlignMA}
Zhiquan Ye, Qian Chen, Wen Wang, and Zhenhua Ling. 2019.
\newblock \href {https://arxiv.org/abs/1908.06725} {Align, mask and select: A
  simple method for incorporating commonsense knowledge into language
  representation models}.
\newblock \emph{ArXiv}, abs/1908.06725.

\bibitem[{Zhao and Vydiswaran(2021)}]{Zhao2021lirex}
Xinyan Zhao and V.G.Vinod Vydiswaran. 2021.
\newblock \href {https://ojs.aaai.org/index.php/AAAI/article/view/17708}
  {Lirex: Augmenting language inference with relevant explanations}.
\newblock \emph{Proceedings of the AAAI Conference on Artificial Intelligence},
  35(16):14532--14539.

\end{thebibliography}
\bibliographystyle{acl_natbib}

\clearpage
\appendix
\begin{table}[]
\centering
\small
\begin{tabular}{@{}llr@{}}
\toprule
\textbf{Dataset} & \textbf{Train} & \textbf{Test} \\ \midrule
\cose v1.11/\ecqa  & 9,741          & 1,221         \\
\quartz           & 2,695          & 783           \\ \bottomrule
\end{tabular}
\vspace{-0.1mm}
\caption{The statistics of 3 datasets in our work.}
\vspace{-0.3cm}
\label{tab:statistics}
\end{table}

\section{Dataset and Implementation Details for Finetuning T5}
\label{app:t5}

Table~\ref{tab:statistics} shows the statistics for the datasets used in our work.

We finetune multiple T5 models~\cite{t5} in our work, and we use HuggingFace~\cite{wolf-etal-2020-transformers} throughout our implementation. We use 512 and 256 for the maximum source length and the maximum target length separately.
To optimize, we use AdamW~\cite{adamw} with a learning rate of 0.0001. We train each model on a NVIDIA RTX 2080 with a batch size of 8 for 30 epochs. 
During inference, we use beam search as the decoding method with a beam size of 2. 
The generation of the EOS token or reaching the maximum target length terminates decoding.

\subsection{Formatting the Seq2Seq Models.}
\label{sec:format}

The formatting of the different models is:
\begin{itemize}[leftmargin=*]
    \item I$\rightarrow$O. Predict the label directly from the question, formatted as: \texttt{context:} \emph{\{question\}} \texttt{options:} \emph{\{concatenated options\}} $\rightarrow$ \emph{\{answer\}}.
    \item IR$\rightarrow$O. Predict the label from the question and the rationale, formatted as: \texttt{context:} \emph{\{question\}} \texttt{options:} \emph{\{concatenated options\}} \texttt{explanation:} \emph{\{rationale\}} $\rightarrow$ \emph{\{answer\}}.
    \item I$\rightarrow$R. Predict the rationale from the question, formatted as: \texttt{explain question:} \{question\} \texttt{answer:} \emph{\{concatenated options\}} $\rightarrow$ \texttt{explanation:} \emph{\{rationale\}}.
\end{itemize}

\subsection{Evaluation of I$\rightarrow \mathbf{R}$ models}
\label{sec:simulatability}
Following \citet{wiegreffe-etal-2021-measuring}, we use \textit{simulatability score} to measure the quality of generated rationales.
Simulatability captures the predictive ability a rationale provides over the input:
\begin{equation}
\small
    \textrm{acc(IR} \rightarrow \textrm{O}) - \textrm{acc(I} \rightarrow \textrm{O}).
\end{equation}

Prior work has shown that simulatability score serves as a reliable measure of rationale quality from the lens of utility to an end user~\citep[\textit{inter alia}]{hase-bansal-2020-evaluating, hase-etal-2020-leakage, rajagopal-etal-2021-selfexplain, PoursabziSangdeh2021ManipulatingAM,wiegreffe-etal-2021-measuring}, and positively correlates with human judgement of rationale utility.
We abstain from reference-based lexical-overlap metrics such as BLEU \cite{papineni-etal-2002-bleu}, which are not suited for measuring plausibility \cite{Camburu2018eSNLINL, Kayser2021eViLAD, clinciu-etal-2021-study}, or faithfulness of rationales \cite{jacovi-goldberg-2020-towards}.

Evaluation via simulability of our generated rationales (\S\ref{sec:generated}) shows negative simulatability for \cose-generated, -13.1 (43.9 - 57.0), and for \ecqa-generated rationales, -12.36 (44.64 - 57.00) where acc (I$\rightarrow$O) = 57.0 (see Tab.~\ref{tab:train-crowdR-comqa}).
This is consistent with findings from \citet{wiegreffe-etal-2021-measuring}.
Perhaps a better evaluation metric for this task is given by leakage-adjusted simulatability \cite{hase-etal-2020-leakage}, where simulatability of non-leaky rationales and those of leaky ones is equally weighted.
We leave a detailed investigation of rationale evaluation to future work.
\section{Human Study Annotation}
\label{app:anno-human-study}
 We recruit workers through our maintained list of qualified workers on Amazon Mechanical Turk (MTurk).
 These workers have collaboration with us on other projects, e.g., AESOP~\cite{sun-etal-2021-aesop}. 
 In addition, we require workers to have completed over 1000 HITs with an approval rate over 99\% and locate in the United States to qualify for our annotation task. 
 As some of the questions contain discussion of adult topics, we warn workers and ask them to terminate the annotation if they are under 18. 
 Our annotation pays for \$1 per HIT. 
 
\begin{figure*}
    \centering
    \includegraphics[width=\textwidth]{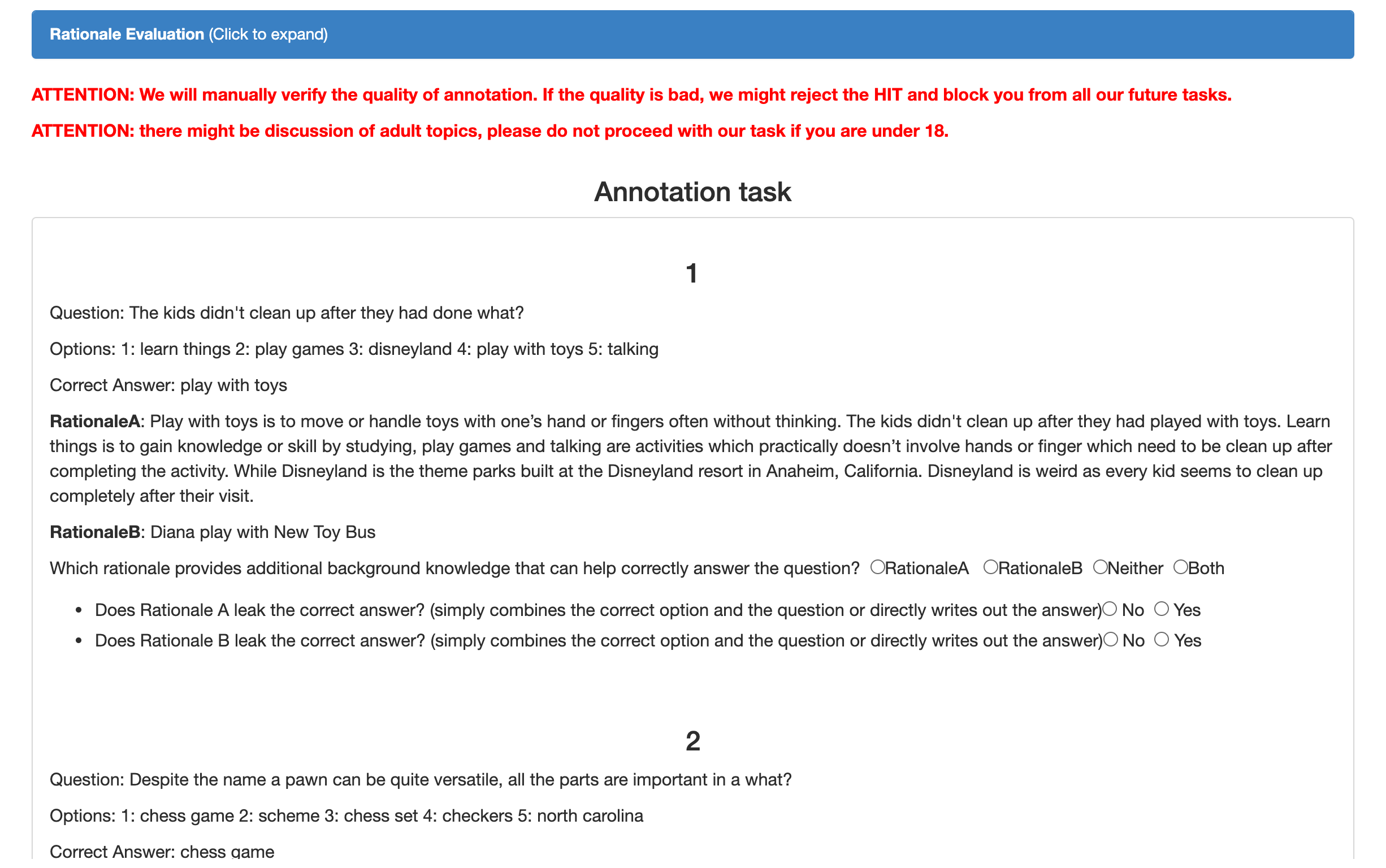}
    \caption{The annotation interface for the direct assessment user study. We use the same interface for both comparing 1) crowdsourced rationales from \cose and \ecqa (\S\ref{sec:human}), as well as 2) crowdsourced v.s. generated rationales from those datasets (\S\ref{sec:generated}). }
    \label{fig:human-direct}
\end{figure*}

\begin{figure*}
    \centering
    \includegraphics[width=\textwidth]{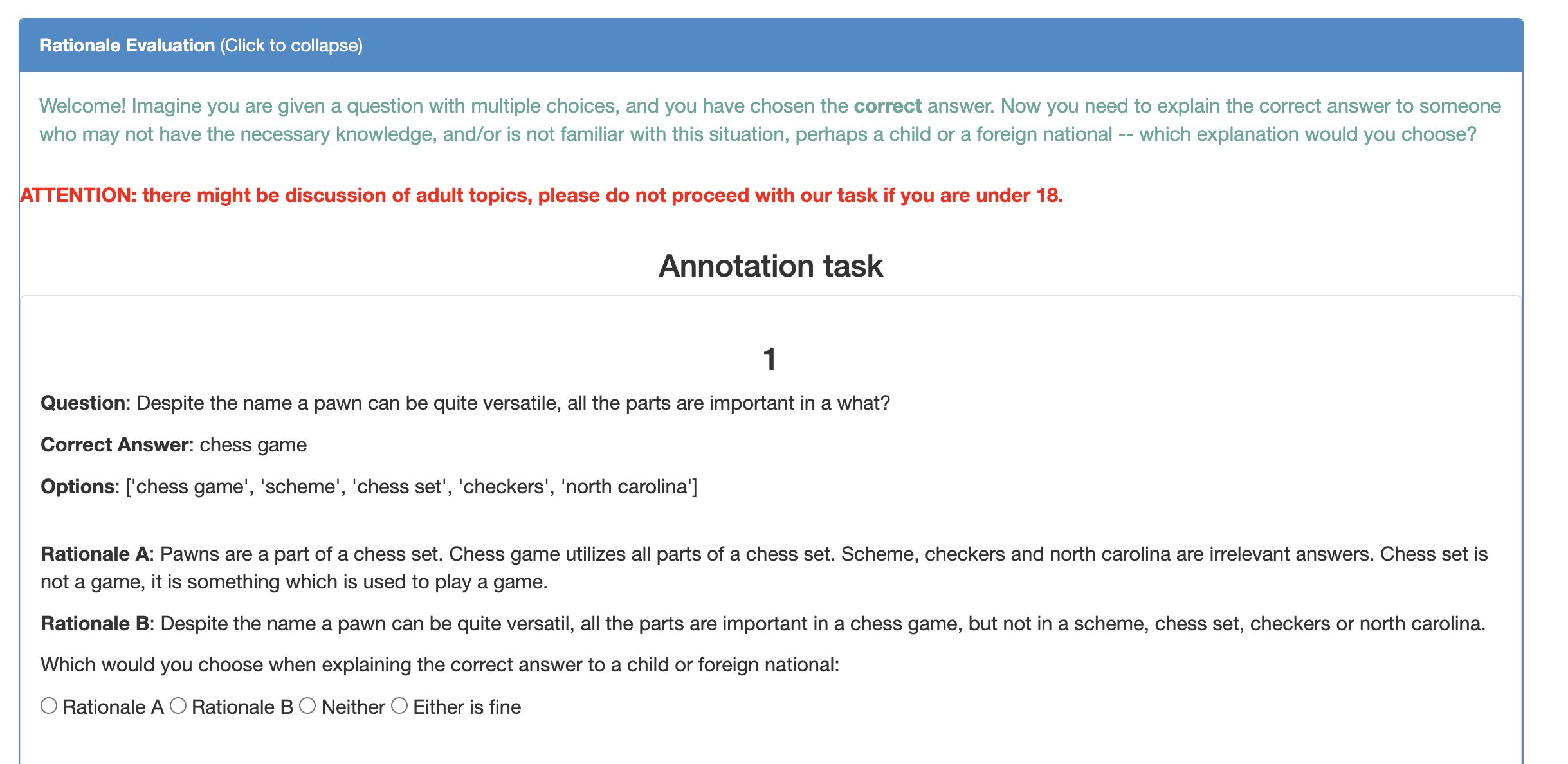}
    \caption{The annotation interface for indirect assessment of if an annotated rationale has background information.}
    \label{fig:human-indirect-background}
\end{figure*}

\begin{figure*}
    \centering
    \includegraphics[width=\textwidth]{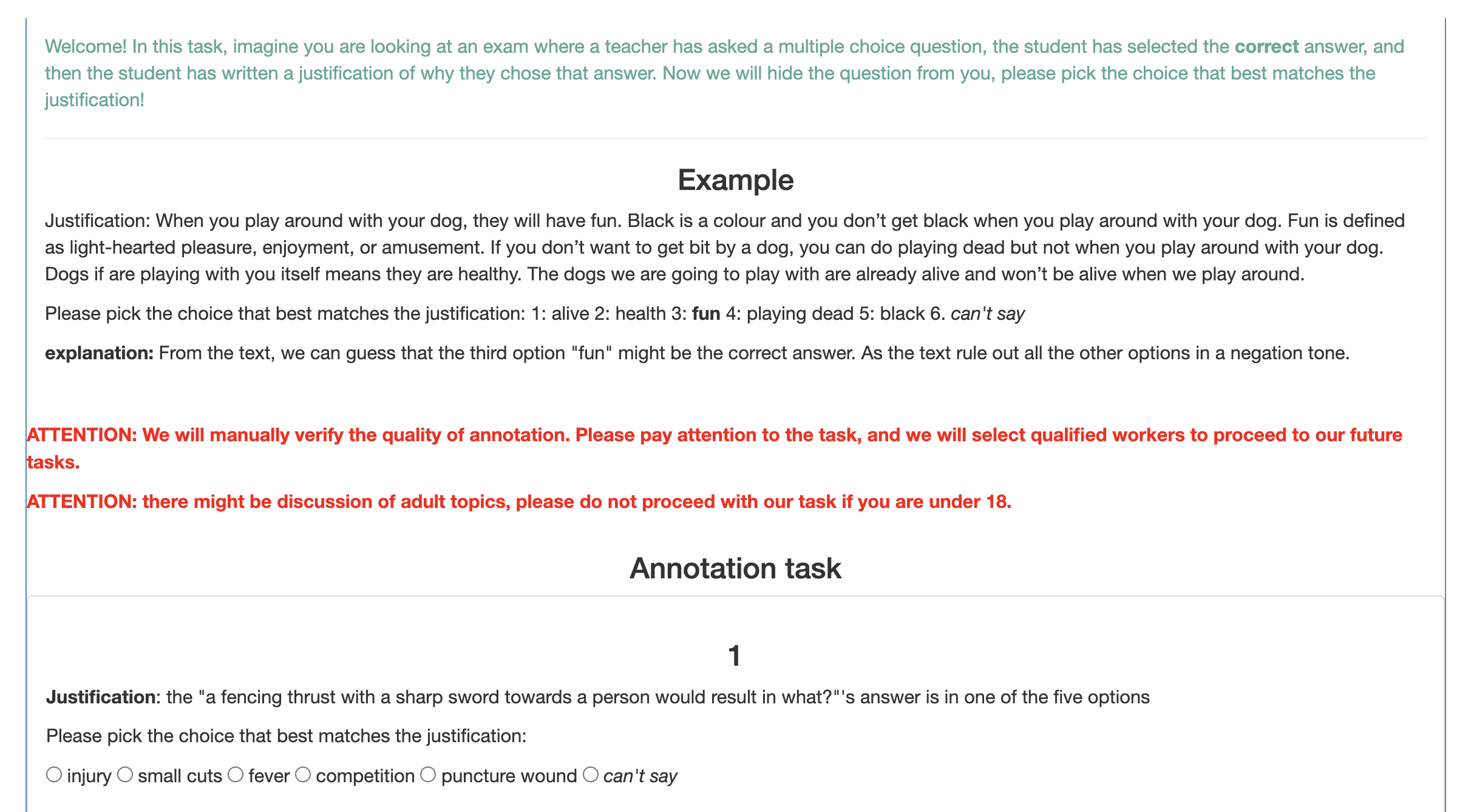}
    \caption{The annotation interface for indirect assessment of if there is leakage in a rationale.}
    \label{fig:human-indirect-leakage}
\end{figure*}

Please see figures \ref{fig:human-direct}, \ref{fig:human-indirect-background}, \ref{fig:human-indirect-leakage} for interfaces to our human studies, discussed in \S\ref{sec:human} and \S\ref{sec:generated}.
 
\section{Author Annotation Details}
\label{app:author}
The 100 CoS-E instances that are annotated by three authors are the same ones annotated by workers on MTurk in the additional quality checks. 
Workers' annotation agrees with ours on 92 out of 100 instances for both background knowledge and leakage annotation, showing a good agreement between us and workers. We do the remaining annotations by ourselves to ensure annotation quality. During the annotation, one example of a \emph{\small$C_{\textrm{no-leak-no-bg}}$} rationale is ``\textit{Rivers flow trough valleys.}'', which occurs in 119 / 1221 instances (9.7\% of the entire dev. set), even though it seemed valid for just one dev. instance. We suspect that this rationale was used as a default placeholder. We urge future researchers to closely inspect the annotation quality before releasing a dataset.



\section{Examples of Generated Rationales}
\label{app:generation-example}

\begin{table*}[]
\small
\resizebox{\textwidth}{!}{
\begin{tabular}{@{}l|l|l|l|l|l@{}}
\toprule
\textbf{Question} & \textbf{Options} & \textbf{Crowdsourced $R_\textrm{\cose}$} & \textbf{Generated $R_\textrm{\cose}$} & \textbf{Crowdsourced $R_\textrm{\ecqa}$} & \textbf{Generated $R_\textrm{\ecqa}$} \\ \midrule
\begin{tabular}[c]{@{}l@{}}What island\\ country is \\ ferret \\ popular?\end{tabular} & \begin{tabular}[c]{@{}l@{}}own home, \\ north carolina, \\ \textbf{great britain}, \\ hutch, \\ outdoors\end{tabular} & \begin{tabular}[c]{@{}l@{}}great britain is \\ the ferret populer\end{tabular} & \begin{tabular}[c]{@{}l@{}}great britain \\ island\end{tabular} & \begin{tabular}[c]{@{}l@{}}Great britain is an island country and Great britain \\ is popular for having ferret as pet. Own home is \\ not island country.  North carolina is not a country \\ but state. Hutch is not country but type of cage. \\ Outdoors is not any country.\end{tabular} & \begin{tabular}[c]{@{}l@{}}Ferrets are not found in hutch. \\ Ferrets are not found in own home \\ or outdoors. Ferret is popular in great \\ britain. Great britain is an island \\ country.\end{tabular} \\ \midrule
\begin{tabular}[c]{@{}l@{}}Where do \\ adults \\ use \\ glue sticks?\end{tabular} & \begin{tabular}[c]{@{}l@{}}classroom, \\ desk drawer,\\ at school, \\\textbf{office}, \\ kitchen drawer\end{tabular} & \begin{tabular}[c]{@{}l@{}}offices employ \\ only Adults.\end{tabular} & \begin{tabular}[c]{@{}l@{}}glue sticks \\ are  used \\ at school\end{tabular} & \begin{tabular}[c]{@{}l@{}}Glue stick is a solid glue used to stick thin paper\\  materials by adults in offices. Adults don't go to \\ classroom and school, and other options don't \\ have adults.\end{tabular} & \begin{tabular}[c]{@{}l@{}}Kids do not use glue sticks in classroom.\\ Desk drawer is a storage space and not a \\ place to store glue sticks. Adults use glue\\ sticks in a classroom. Office is a room or \\ set of rooms or a building where adults\\  work, usually sitting at desks. Kitchen \\ drawer is a storage space and not a place \\ for adults to use glue sticks. School is a \\ place for learning and not a place to use\\  glue sticks.\end{tabular} \\ \midrule
\begin{tabular}[c]{@{}l@{}}What could \\ go \\ on top of \\ wood?\end{tabular} & \begin{tabular}[c]{@{}l@{}}lumberyard, \\ synagogue. \\ floor,\\\textbf{carpet}, \\ hardware store\end{tabular} & Carpets in Chennai & \begin{tabular}[c]{@{}l@{}}carpets are \\ made \\ of wood\end{tabular} & \begin{tabular}[c]{@{}l@{}}It is the carpet that could go on top of wood. \\ Lumberyard is a place that cant go anywhere. \\ Floor is at the bottom. Synagogue is a building\\  that could not go on the top of wood. A hardware\\  store can never go on the top of the wood.\end{tabular} & \begin{tabular}[c]{@{}l@{}}Flooring cant go on top of wood. \\ Hardware\\ store or lumberyard or synagogue is not\\ made of wood. Carpet can be made of \\ wood.\end{tabular} \\ \midrule
\begin{tabular}[c]{@{}l@{}}Where is a \\ doormat \\ likely to be \\ in front of?\end{tabular} & \begin{tabular}[c]{@{}l@{}}facade, \\\textbf{front door}, \\ doorway, \\ entrance porch, \\ hallway\end{tabular} & \begin{tabular}[c]{@{}l@{}}Your front door \\ makes an entrance\end{tabular} & \begin{tabular}[c]{@{}l@{}}a doormat \\ likely \\ to be \\ in front of \\ entrance\end{tabular} & \begin{tabular}[c]{@{}l@{}}Front door is the main entrance to a house. There\\ is a doormat likely to be in front of the front door. \\ Façade is the principal front and keeping doormat\\ there doesn’t solve the issue. Doorway is the \\ entrance to room through a door and not a place \\ where a doormat is likely to be placed. Entrance porch\\ is a covered shelter in front of the entrance and \\ doormat is not likely to be here. Hallway is not \\ likely to have a doormat for wiping feet.\end{tabular} & \begin{tabular}[c]{@{}l@{}}A doormat is likely to be in front of \\ front door. Doorway is the entrance \\ to a room through a door. Doormat \\ cant be found in front of entrance \\ porch or hallway. Façade is not a \\ place where a doormat can be \\ found.\end{tabular} \\  \midrule
\begin{tabular}[c]{@{}l@{}}The teacher \\ doesn't \\ tolerate \\ noise during \\ a test in \\ their what?\end{tabular} & \begin{tabular}[c]{@{}l@{}}movie theatre, \\ bowling alley, \\ factory, \\ store, \\ \textbf{classrooom}\end{tabular} & \begin{tabular}[c]{@{}l@{}}Rivers flow trough \\ valleys.\end{tabular} & \begin{tabular}[c]{@{}l@{}}a classroom \\ is a \\learning place\end{tabular} & \begin{tabular}[c]{@{}l@{}}A teacher takes the test in a classroom where she \\ does not tolerate any noise. All the other options \\ are incorrect as a test is not conducted\\ in any of the options.\end{tabular} & \begin{tabular}[c]{@{}l@{}}A classroom is a place where one sits \\ and takes test. All the other options are \\ incorrect as they are not a place\\ where one can sit and take test. \end{tabular} \\  \midrule
\begin{tabular}[c]{@{}l@{}}If you \\ have to\\ read a\\ book that \\is very dry \\ and long \\ you may \\become what?\end{tabular} & \begin{tabular}[c]{@{}l@{}}have time, \\ boring, \\ learn new, \\ enjoyable, \\ \textbf{bored}\end{tabular} & \begin{tabular}[c]{@{}l@{}}a book is world.\end{tabular} & \begin{tabular}[c]{@{}l@{}}boring is \\ when you \\ have to read\\ a book\end{tabular} & \begin{tabular}[c]{@{}l@{}}You may get bored while reading a long and dry \\ book. While reading a long book, you dont have time.\\ A book that seems to be very dry is \\ boring. You cant become learn new. A book which is\\ very dry and long can never be enjoyable.\end{tabular} & \begin{tabular}[c]{@{}l@{}}Dry and long books make us feel bored. \\Boring comes under having time\\ and the other options\\ are not caused by reading dry and \\ long books. So if you have to read \\a book that is very dry and long \\you may become bored. \end{tabular} \\  \midrule
\begin{tabular}[c]{@{}l@{}}If you are\\ awaking multiple\\ times throughout \\the night\\ because a lot\\ is on your\\ mind, what\\ is a \\likely cause?\end{tabular} & \begin{tabular}[c]{@{}l@{}}irritability, \\ \textbf{depression}, \\ getting out of bed, \\ happiness, \\ discomfort\end{tabular} & \begin{tabular}[c]{@{}l@{}}Depression indicates\\ mental distress.\end{tabular} & \begin{tabular}[c]{@{}l@{}}mental or\\ emotional states\end{tabular} & \begin{tabular}[c]{@{}l@{}}Depression is an illness of intense sadness and \\having a lot in mind while not letting one to sleep is\\ likely caused by depression. The other options \\do not cause one to have a lot in mind.\end{tabular} & \begin{tabular}[c]{@{}l@{}}Depression is not caused by \\ waking up  multiple times  throughout \\ the night. Getting out of bed \\ is not a cause. Happiness \\ cant be caused by  waking up \\ multiple times  throughout the night.\\ By waking multiple times throughout\\ the night, a lot of things are on your \\  mind.Discomfort is a likely cause of\\  irritability. \end{tabular} \\\midrule
\begin{tabular}[c]{@{}l@{}}What do \\you want\\ someone to\\ do when\\ you illustrate\\ point?\end{tabular} & \begin{tabular}[c]{@{}l@{}}did not understand, \\ accepting, \\ make clear, \\ understood, \\ \textbf{understanding}\end{tabular} & \begin{tabular}[c]{@{}l@{}}we need a\\ understanding.\end{tabular} & \begin{tabular}[c]{@{}l@{}}make clear \\what do you\\ want someone\\ to do\\ when you illustrate\end{tabular} & \begin{tabular}[c]{@{}l@{}}To illustrate is to make something clearer and more visible.\\ We want someone to understand when we \\illustrate point. Did not understand being opposite of \\what you want someone to do,  accepting and \\Make clear does not relate to and while\\ understood is a past tense.\end{tabular} & \begin{tabular}[c]{@{}l@{}}When someone illustrates point, they want \\ to make clear about it. \\To understand something is not appropriate \\ here. If someone did not understand the \\point illustrated then he will not understand\\ it. Someone cant accept or understand\\ the point if he illustrates it. \end{tabular} \\

\bottomrule
\end{tabular}}
\caption{Examples of crowdsourced and generated rationales from \cose v1.11 and \ecqa. 
The ground truth options are \textbf{in bold}.
As can be clearly seen, the crowdsourced \cose rationales are often ungrammatical, and off-topic, and do not provide the background knowledge necessary to understand the ground truth answers.
In contrast, crowdsourced \ecqa rationales are grammatical, and provide the necessary background knowledge for human interpretability.
Moreover, generated rationales are often factually incorrect, such as ``\emph{carpets are made of wood}'', and also lack much of the commonsense reasoning necessary for rationales.
All types of rationales leak the correct answer.
}
\label{tab:qualitative}
\end{table*}

Table~\ref{tab:qualitative} show examples of generated rationales for \cose and \ecqa, together with crowdsourced rationales from the original datasets. 
Based on our human evaluation, the quality of generated rationales is worse than that of the crowdsourced rationales; hence, generated rationales provide weaker signals for training IR$\rightarrow$O models.

\end{document}